\documentclass[twoside]{article}

\usepackage[preprint]{aistats2025}
\usepackage[round]{natbib}
\renewcommand{\bibname}{References}
\renewcommand{\bibsection}{\subsubsection*{\bibname}}

\defcitealias{mosaic_consortium_mosaic_2025}{MOSAIC consortium (2025)}
\defcitealias{mosaic_consortium_mosaic_2025_p}{(MOSAIC consortium 2025)}
\defcitealias{the_cancer_genome_atlas_program_cancer_2022}{TCGA (2012)}
\defcitealias{the_cancer_genome_atlas_program_cancer_2022_p}{(TCGA 2012)}

\usepackage[utf8]{inputenc} % allow utf-8 input
\usepackage[T1]{fontenc}    % use 8-bit T1 fonts
\usepackage{url}            % simple URL typesetting
\usepackage{booktabs}       % professional-quality tables
\usepackage{amsfonts}       % blackboard math symbols
\usepackage{nicefrac}       % compact symbols for 1/2, etc.
\usepackage{microtype}      % microtypography
\usepackage{xcolor}         % colors
\usepackage{amsmath}
\usepackage{hyperref}
\usepackage[capitalize,noabbrev]{cleveref}
\usepackage{amssymb}
\usepackage{mathtools}
\usepackage{mathrsfs}
\usepackage{amsthm}
\usepackage{dsfont}
\usepackage{xtab}
\usepackage{subcaption}
\usepackage{xfrac}
\usepackage{tikz}
\usepackage{listings}
\usepackage{cuted}

\usepackage{booktabs}
\usepackage{tabularx}
\usepackage{ragged2e}

\usepackage{longtable}
\usepackage{array}

\usepackage[acronym]{glossaries-extra}
\glsdisablehyper
\setabbreviationstyle[acronym]{long-short}

\newacronym{rl}{RL}{Reinforcement Learning}
\newacronym{rlvr}{RLVR}{Reinforcement Learning from Verifiable Rewards}
\newacronym{sft}{SFT}{Supervised Fine-Tuning}
\newacronym{cot}{CoT}{Chain of Thought}
\newacronym{llm}{LLM}{Large Language Model}
\newacronym{grpo}{GRPO}{Group Reward Policy Optimization}
\newacronym{qa}{Q\&A}{Question and Answer}
\newacronym{dea}{DEA}{Differential Expression Analysis}
\newacronym{spde}{SpDE}{Spatial Differential Expression}
\newacronym{tvhe}{TvHE}{Tumour vs Healthy Expression}
\newacronym{gi}{GI}{Gene Indication Features}
\newacronym{tcgasa}{TCGASA}{TCGA Signature Activity}
\newacronym{dseqde}{DSeqDE}{DrugSeq Differential Expression Analysis}
\newacronym{dpp}{DPP}{Drug-Pathway Perturbation}
\newacronym{ttp}{TTP}{Therapeutic Target Profiling}
\newacronym{sd}{SD}{Structural Druggability}

\makeglossaries

\theoremstyle{plain}

\theoremstyle{definition}

\theoremstyle{remark}

\DeclareRobustCommand*\heavyplus{\tikz[line width=0.8ex, scale=0.3, baseline=-0.15ex]\draw(0,0.5)--(1,0.5)(0.5,0)--(0.5,1);}
\DeclareRobustCommand*\heavycross{\tikz[line width=1ex, scale=0.3, baseline=-0.15ex]\draw(0,0)--(1,1)(0,1)--(1,0);}

\newcommand\blfootnote[1]{%
  \begin{NoHyper}%
  \renewcommand\thefootnote{}\footnote{#1}%
  \addtocounter{footnote}{-1}%
  \end{NoHyper}%
}

\begin{document}


\runningauthor{Bigaud N., Cabeli V., G\"urel M., Pignet A., Klein J., Wainrib G., Durand E.}
\twocolumn[
\aistatstitle{OwkinZero: Accelerating Biological Discovery with AI} 
\aistatsauthor{Nathan Bigaud\textsuperscript{*}  \\ \And Vincent Cabeli\textsuperscript{*} \\ \And Meltem G\"{u}rel\textsuperscript{*} \\ \And Arthur Pignet\textsuperscript{*} \\ \AND John Klein\textdagger \And Gilles Wainrib\textdagger \And Eric Durand\textdagger \AND}

\aistatsaddress{Owkin}
]
\blfootnote{\textsuperscript{*} Core team, equal contribution, alphabetical order.}
\blfootnote{\textdagger Senior authors.}

\begin{abstract}
    While large language models (LLMs) are rapidly advancing scientific research, they continue to struggle with core biological reasoning tasks essential for translational and biomedical discovery. To address this limitation, we created and curated eight comprehensive benchmark datasets comprising over 300,000 verifiable question-and-answer pairs, each targeting critical challenges in drug discovery including target druggability, modality suitability, and drug perturbation effects. 
    %We release a first-of-its-kind benchmark to enable systematic assessment of LLM performance in biological reasoning tasks.
Using this resource, we developed the OwkinZero models by post-training open-source LLMs through a Reinforcement Learning from Verifiable Rewards strategy. Our results demonstrate that specialized 8--32B OwkinZero models substantially outperform larger, state-of-the-art commercial LLMs on these biological benchmarks. Remarkably, we uncover evidence of a key aspect of generalization: specialist models trained on a single task consistently outperform their base models on previously unseen tasks. This generalization effect is further amplified in our comprehensive OwkinZero models, which were trained on a mixture of datasets and achieve even broader cross-task improvements.
This study represents a significant step toward addressing the biological reasoning blind spot in current LLMs, demonstrating that targeted reinforcement learning on carefully curated data can unlock generalizable performance in specialized models, thereby accelerating AI-driven biological discovery.

\end{abstract}

\section{INTRODUCTION}
\Glspl{llm} are rapidly improving at multi-step reasoning, creating an opportunity to accelerate scientific discovery~\citep{wang_scientific_2023, gottweis_towards_2025}.
Biology is a natural proving ground for these emerging capabilities: it spans multiple scales of organisation, relies on diverse data modalities, and demands careful integration of mechanism with evidence.
Yet, despite impressive general performance, current state-of-the-art \Glspl{llm} remain subpar on specialised biological tasks that require reasoning over expression patterns, perturbation effects, structural constraints, and therapeutic priors~\citep{wang_txgemma_2025, laurent_lab-bench_2024, zhao_biomaze_2025}.

To address this gap, we turn to the \gls{rlvr} paradigm introduced in~\citep{deepseek-ai_deepseek-r1_2025} to post-train open models directly on verifiable biological questions.
Biology is first and foremost an experimental science: new knowledge is obtained through experimentation rather than deduction, usually via high-throughput screening assays, clinical trials, or population studies~\citep{marx_big_2013}.
As a result, valid reasoning traces can be diverse, largely unknown, and rarely annotated.
\gls{rlvr} is uniquely suited to this problem as it optimizes for answer accuracy directly, removing the need for ground-truth \gls{cot}.

Building on this methodology's success in domains like mathematics, code generation, and logic puzzles, we applied it to a curated suite of biology-first tasks that reflect key steps in the drug discovery pipeline. Concretely, we generate a collection comprising over 300{,}000 \gls{qa} pairs across the following task families: Tumour vs Healthy tissue Differential Expression (including at fine spatial resolution), Signature Activity Comparison across cancer types, Drug Perturbation Effect Prediction, Therapeutic Target Profiling, and Structural Druggability.

We find that specialised models as small as 8 to 32B parameters, trained with a single \gls{rl} phase, achieve state-of-the-art accuracy across in-domain biological tasks and, in several cases, exhibit meaningful out-of-domain generalisation -- surpassing larger commercial models on our benchmarks.
Beyond accuracy, we analyse reasoning quality and faithfulness: single-task \gls{rl} improves answer accuracy and consistency between reasoning and answers on their respective domains, while mixture training increases overall accuracy with weaker faithfulness, motivating additional alignment stages.

The remainder of the paper introduces the OwkinZero framework, details dataset construction and verifiers, reports comprehensive benchmarking against strong baselines and SOTA models, and discusses implications for building specialised scientific reasoners.

Our work represents, to the best of our knowledge, the first explicit attempt to create a language model with deep biological reasoning abilities through the use of \gls{rlvr} on a set of biological tasks. We summarize our contributions as follows:

\begin{itemize}
    \item We introduce a new benchmark of eight datasets with over 300,000 verifiable \gls{qa} pairs, designed to test complex problem-solving across the drug discovery pipeline.

    \item We demonstrate that specialized models, post-trained via reinforcement learning, substantially outperform larger, state-of-the-art commercial \glspl{llm} on our biological benchmarks.

    \item We uncover insights into cross-task generalization, where specialist models trained on a single task show improved performance on unseen, out-of-domain tasks compared to their base models.

    \item Our OwkinZero models, trained on a mixture of datasets, amplify this effect, achieving broader cross-task generalization and outperforming single-task specialists even on their respective in-domain tasks.
\end{itemize}

\section{DATASETS}

To enable the post-training of \glspl{llm} via \gls{rl} for improved biological reasoning, we generated a collection of datasets together comprising over 300,000 \gls{qa} pairs.
These datasets encompass a breadth of biological domains and analytical techniques, including transcriptomics, perturbation assays, molecular signatures, druggability assessment, and structural biology.

One of our main considerations was to ensure the integrity of our evaluation and mitigate the risk of pre-training data contamination, which could lead to performance gains from better recall rather than better reasoning.
Where feasible, we prioritised the inclusion of proprietary and newly published source data released after the pre-training cutoff of the \glspl{llm} we employed, such as the MOSAIC (Multi-Omics Spatial Atlas in Cancer) dataset ~\citetalias{mosaic_consortium_mosaic_2025_p}, and the Tahoe-100M dataset~\citep{zhang_tahoe-100m_2025}.
When drawing from older publicly accessible sources such as The Cancer Genome Atlas~\citetalias{the_cancer_genome_atlas_program_cancer_2022_p}, we ensured that the structured \gls{qa} pairs offered a level of processing and contextual richness well beyond what the models were likely exposed to during their pre-training.
We subjected each dataset to a conservative, task-specific train/test split, ensuring that no subject entity (e.g., gene, cancer indication, or drug) in the test set was shared with the training set.

Transforming raw biological data into verifiable questions that test reasoning is a non-trivial step.
Our approach was an iterative process involving significant expert curation: domain experts were involved in the design of the question templates, and benchmarking was used to tune the difficulty and understand what current models are able to solve.
Following these principles, all datasets were systematically formulated into a standardised natural language multiple-choice \gls{qa} format.
The questions were deliberately crafted to be solvable through logical inference but pose substantial reasoning challenges, requiring the integration of the provided context with background biological knowledge rather than simple information recall.
The computational methods used to generate these \gls{qa} pairs included \gls{dea} and single-sample gene set enrichment analysis (ssGSEA)~\citep{barbie_ssgsea_2009} applied to diverse data sources such as TCGA and Tahoe-100M, and spatial analyses performed on MOSAIC data.

An overview of the datasets, including their task formulation, sample sizes, and associated source domains, is presented in~\cref{tab:datasets}.
Further dataset details, including the curation process, \gls{qa} schema along with example \gls{qa} pairs, and train/test split strategy, are available in~\cref{sec:suppdatasets}.

\subsection{Expression-Based Datasets}

\paragraph{\glsfirst{spde}}:
Derived from in-house feature scores obtained from MOSAIC Visium spatial transcriptomics data, this dataset probes gene-level contrasts between \emph{tumour islets} and \emph{stroma} at indication level. Regions are obtained via spatial deconvolution and spatial processing to label tumour islets vs.\ stromal compartments. For each $(\text{$indication$}, \text{$gene$})$ we compute a spatial differential-expression score for the tumour-islet vs.\ stroma contrast. Question items are binary A/B prompts that ask which of two genes significantly exhibits the specified direction of change (\emph{upregulated in tumour islets relative to stroma} or \emph{downregulated in tumour islets relative to stroma}). Positives are sampled from extreme tails (e.g., $s \ge Q_{0.99}(s)$ for upregulated, $s \le Q_{0.01}(s)$ for downregulated, where $s$ denotes the in-house spatial contrast feature score and \(Q_p\) denotes the \(p\)-quantile), while distractors are drawn from the same indication with values outside the extreme tail for the queried direction (upregulated distractors: $s \le 0.5$; downregulated distractors: $s \ge -0.5$). Answers (A/B) are randomized per item. Train/test splits are disjoint in $(\text{$indication$}, \text{$gene$})$, stratified by $indication$ and question type (up/down) to preserve class balance.

This task integrates crucial biological knowledge for identifying potential therapeutic targets by comparing two functionally distinct regions of the tumour microenvironment~\citep{jin_advances_2024, liu_identification_2023}.
By sourcing this data from the contemporary MOSAIC cohort, it is specifically designed to uncover patterns relevant to unmet medical needs in patients under the current standard of care.

\paragraph{\glsfirst{tvhe}}:
This dataset contains questions comparing transcript abundance of genes in tumour versus adjacent normal tissues across various TCGA cancer types. Each question targets a (\textit{gene}, \textit{indication}) pair and asks which tissue type shows higher expression.
For each $indication$ with sufficient adjacent normal samples, we perform a Wilcoxon rank-sum test (using \texttt{rank\_genes\_groups} from \texttt{scanpy}~\citep{scanpy2018}) to compare gene expression between tumour and adjacent normal samples, correcting $p$-values by Benjamini-Hochberg FDR. We retain genes meeting below thresholds to form our \gls{qa} pairs:
\[
\begin{aligned}
\text{Tumour-up:}\quad & \mathrm{FDR}<0.05 \ \wedge\ \log_2\mathrm{FC}>1, \\
\text{Normal-up:}\quad & \mathrm{FDR}<0.05 \ \wedge\ \log_2\mathrm{FC}<-1.
\end{aligned}
\]

Each \gls{qa} item originally asks which tissue shows higher expression for $(\text{$indication$}, \text{$gene$})$, with options \texttt{tumour tissue} and \texttt{adjacent normal tissue}. However, we rephrase each \gls{qa} pair systematically from these original simple formulations to incorporate formal biomedical language such as ``transcript abundance'', ``neoplastic cells'', and ``non-neoplastic tissue'', without altering the underlying structures. To give an example, this dataset asks: ``In Esophageal carcinoma, does VAC14-AS1 exhibit elevated transcript levels in the neoplastic cells compared to the surrounding non-neoplastic tissue?'' with the answer options ``A) neoplastic tissue'' and  ``B) non-neoplastic tissue''. Each original simple-form \gls{qa} pair is mapped to exactly one rephrased version, ensuring a strict one-to-one correspondence and preserving the total number of questions in the dataset. Splits are disjoint in $(\text{$gene$}, \text{$indication$})$ and stratified by direction and $indication$. Metadata captures these rewording strategies supporting research into natural language variation, biomedical comprehension, and expression-based reasoning.
Much like the \glsxtrshort{spde} dataset, this dataset represents a standard analysis of patient data to identify potential drivers of tumour pathology or actors in the immune response.

\paragraph{\glsfirst{gi}}:
Comprising 127,069 training examples, the \glsxtrshort{gi} dataset presents \emph{True/False} statements tied to $(\text{$gene$}, \text{$indication$})$ pairs derived from in-house MOSAIC single-cell/spatial feature scores, covering orthogonal feature families such as: indication-level contrasts vs.\ reference tissues (e.g., blood, spleen, liver), malignant vs.\ non-malignant and malignant vs.\ stromal enrichment, intra-tumour heterogeneity across malignant subpopulations, genomic alteration frequencies (e.g., copy-number events). Similar to \glsxtrshort{tvhe}, rather than simple statements, the questions are deliberately reformulated to introduce more technical phrasing and clinical precision, for example, substituting ``higher expression'' with ``significantly elevated transcript abundance'' or ``low heterogeneity'' with ``minimal variability across malignant subpopulations''. Again as with the \glsxtrshort{tvhe} dataset, for every simple-form \gls{qa} pair, we create exactly one rephrased counterpart, selected from the rewording strategies. This guarantees a one-to-one mapping, so the overall question count remains unchanged. These variations test the ability to reason across different linguistic framings while maintaining a consistent \gls{qa} structure. Splits are disjoint in $(\text{$gene$}, \text{$indication$})$ and stratified by question/feature type and label.

This dataset brings together diverse biological signals, ranging from spatial topology and malignant heterogeneity to genomic alteration patterns, uniformly cast into a verifiable True/False format. By consolidating these orthogonal feature types into a single $(gene, indication)$-indexed task with controlled linguistic variation, it offers a dense, information-rich setting for evaluating model performance.

\clearpage
\onecolumn
\begingroup
\renewcommand{\arraystretch}{2}
\setlength{\tabcolsep}{3pt}

\begin{longtable}{%
  >{\raggedright\arraybackslash}p{0.122\linewidth}% Dataset
  >{\raggedright\arraybackslash}p{0.335\linewidth}% Description
  >{\raggedright\arraybackslash}p{0.160\linewidth}% Task
  >{\centering\arraybackslash} p{0.145\linewidth}% Size
  >{\raggedright\arraybackslash}p{0.160\linewidth}% Source
}
\toprule
\textbf{Dataset} & \textbf{Description} & \textbf{Task} & \textbf{Size (Train/Test)} & \textbf{Source} \\
\midrule
\endfirsthead

\multicolumn{5}{c}{{\bfseries Continued from previous page}} \\
\toprule
\textbf{Dataset} & \textbf{Description} & \textbf{Task} & \textbf{Size (Train/Test)} & \textbf{Source} \\
\midrule
\endhead

\midrule
\multicolumn{5}{r}{{Continued on next page}} \\
\endfoot

\bottomrule
\caption{Dataset summary illustrating the diversity of data modalities and tasks encompassed by the collection. Each row corresponds to a distinct dataset. Columns indicate: \textbf{Dataset} (short name and acronym), \textbf{Description} (brief summary of dataset scope and biological context), \textbf{Task} (type of domain task), \textbf{Size (Train/Test)} (number of \gls{qa} pairs in the training and test splits), and \textbf{Source} (origin of the underlying data). 
\label{tab:datasets}}\\
\endlastfoot

\glsfirst{spde} &
Questions identifying up/downregulated genes in tumour microenvironments. &
Spatial transcriptomics &
1,092 / 81 &
MOSAIC \\

\glsfirst{tvhe} &
Comparing gene abundance in neoplastic versus corresponding non-neoplastic tissue. &
(Differential) gene expression &
49,488 / 1,758 &
TCGA\\

\glsfirst{gi} &
Single-cell and spatial transcriptomics question set spanning multiple biological axes. &
Spatial transcriptomics &
127,069 / 22,484 &
MOSAIC\\

\glsfirst{tcgasa} &
Signature-based question set comparing gene activity both within and across cancers. &
Signature-based expression &
35,969 / 1,600 &
TCGA, DSigdb \\

\glsfirst{dseqde} &
Perturbation-based questions predicting gene deregulation from drug-target interactions. &
Drug-target perturbation analysis &
23,169 / 2,731 &
Perturbation assays\\

\glsfirst{dpp} &
Identification of pathways most perturbed by drug treatment, using enrichment scores. &
Drug-pathway perturbation analysis &
8,000 / 2,000 &
Tahoe-100M, Reactome\\

\glsfirst{ttp} &
Multi-domain questions on druggability, modality, safety, and disease relevance of genes. &
Target druggability &
2,482 / 276 &
Uniprot, patents databases, CT databases\\

\glsfirst{sd} &
Structural comparison of protein binding sites to assess druggability. &
Structural biology &
7,141 / 376 &
TOUGH-M1\\
\end{longtable}
\endgroup
\clearpage
\twocolumn

\subsection{Signature-Based Datasets}

\paragraph{\glsfirst{tcgasa}}:
This dataset combines multiple question types centered on signature-based expression reasoning across TCGA cancer types.
Questions ask about relative expression of gene signatures, similarity of signature activity distributions, and cancer-type similarities inferred from transcriptomic profiles.
Example formats include: “Which cancer type has higher expression of the l-thyroxine signature?”, “Which signature has a more similar distribution to ethinyl estradiol across all cancer types?”, and “In Kidney Chromophobe, which signature has higher expression?”.
Signatures are derived from the DSigDB~\citep{yoo_dsigdb_2015} database, and correspond to the list of differentially expressed genes after being treated with a given drug.
In the questions, signatures are described with the corresponding drug name as well as up to 10 random genes composing the signature.
All questions follow a multiple-choice format and are phrased using formal biomedical language.

This dataset aims to mimic standard methods in drug discovery.
Comparing pathway-level activities across cohorts of different cancer types, for example, is often used as a first approach in drug repositioning efforts as similar transcriptomic distributions can be used to infer similar biological pathways activations.

\subsection{Drug Effects and Perturbation Studies}

\paragraph{\glsfirst{dseqde}}:
This dataset leverages proprietary drug perturbation assays to test whether inhibition of a specific drug target would lead to transcriptional changes in cancer cells. Compound metadata are curated to map each treatment to inhibitory $target$ annotations; only inhibitors/antagonists/degraders are retained to preserve a loss-of-function interpretation. For each $(\text{$target$}, \text{context})$, differential expression contrasts (treated vs.\ control) provide sets of deregulated genes.

We instantiate three item types:
\begin{enumerate}
  \item \textbf{Yes/No (gene-level):} ``Would inhibiting $target$ deregulate $gene$ in $indication$ cells?''
  \item \textbf{Pairwise (gene-level):} ``Which gene ($A$ or $B$) is deregulated by inhibiting $target$ in $indication$ cells?''
  \item \textbf{Pairwise (pathway-level):} map DEGs to Reactome~\citep{milacic_reactome_2024}; ask ``Which Reactome $pathway$ would be deregulated by a drug inhibiting the activity of a $target$ in $indication$ cells?''
\end{enumerate}
Splits are \emph{entity-disjoint}: no overlap in $target$, any $gene$ appearing in stem or options, or $pathway$ options between train and test; A/B ordering is randomized independently per split.

\paragraph{\glsfirst{dpp}}:
Derived from perturbation-response profiles in Tahoe-100M, this dataset asks for the \emph{most perturbed Reactome pathway} (with direction) under a specified $(\text{$drug$}, \text{$cell\_line$}, \text{$concentration$})$ context. For example: ``Which Reactome gene set would be most significantly affected by Bimatoprost at 0.05 µM in SW1417 cells, and in which direction: upregulation or downregulation?''
For each context, we compute treated vs.\ control differential ranks and run ssGSEA over Reactome gene sets, keeping pathways with FDR\,$<0.05$. Questions are formatted as multiple-choice with paired pathway-direction options: we select the one with maximal absolute normalized enrichment score ($|\mathrm{NES}|$) as the correct answer while picking a distractor from either the full pathway space or from the retained context-specific enriched pathways (FDR $<0.05$) allowing for levels of difficulty which are recorded in the metadata of each \gls{qa} item. The train/test split enforces: (i) no $drug$ overlap; (ii) no $cell\_line$ overlap; (iii) Reactome modules are disjoint by assigning non-overlapping ontology subtrees to train/test, further filtering test sets with maximum train–test Jaccard similarity $\le 0.3$. This prevents pathway leakage via near-duplicate terms and ensures robust evaluation.

Such potential outcomes puzzles are the gold standard for evaluating the ability of a model to reason about the biological consequences of drug perturbations, and answering these type of questions is seen as a key step towards future drug development~\citep{noutahi_virtual_2025, bunne_how_2024}.

\subsection{Target Druggability and Structural Assessment}

\paragraph{\glsfirst{ttp}}:
This dataset spans 28 distinct question types designed to evaluate the therapeutic potential of targets across diverse biological and pharmacological dimensions.
Topics include antibody and small molecule tractability, structural characterization, toxicity and safety concerns, ligandability, and disease relevance such as cancer or inflammatory conditions.
Questions are expressed through a variety of natural language templates, ranging from direct assertions to more exploratory or hedged phrasings (e.g., “Can PRDX5 be targeted by antibodies?”, “Is it true that HEG1 is druggable with monoclonal antibodies?”, “Is TIGIT associated with cancer pathways?”).
Built from structured annotations aggregated from UniProt, patent literature, and clinical trial databases, this dataset enables comprehensive reasoning over target viability.

\paragraph{\glsfirst{sd}}:
Focused on identifying the more druggable of two candidate binding sites within a protein, this dataset presents pairwise comparison questions grounded in structural data.
Binding sites (pockets) and their druggability scores are computed using Fpocket~\citep{le_guilloux_fpocket_2009} on all protein structures sourced from the TOUGH-M1~\citep{tough-m1_govindaraj_2018} dataset. An example question from this dataset reads: ``Given the protein with amino-acid sequence (provided as a sequence of residues, no 3D coordinates), which one of these two binding sites (presented in the form of their corresponding list of residues with respect to the original sequence) has the highest druggability score?''.
The dataset supports structure-informed assessment of protein druggability using sequence-defined input, enabling evaluation of binding site quality across diverse protein targets.

\section{RESULTS}

% add small introduction to the results

\subsection{Multi Task Biological Benchmarking}

\begin{figure*}[h]
    \centering
    \includegraphics[width=\textwidth]{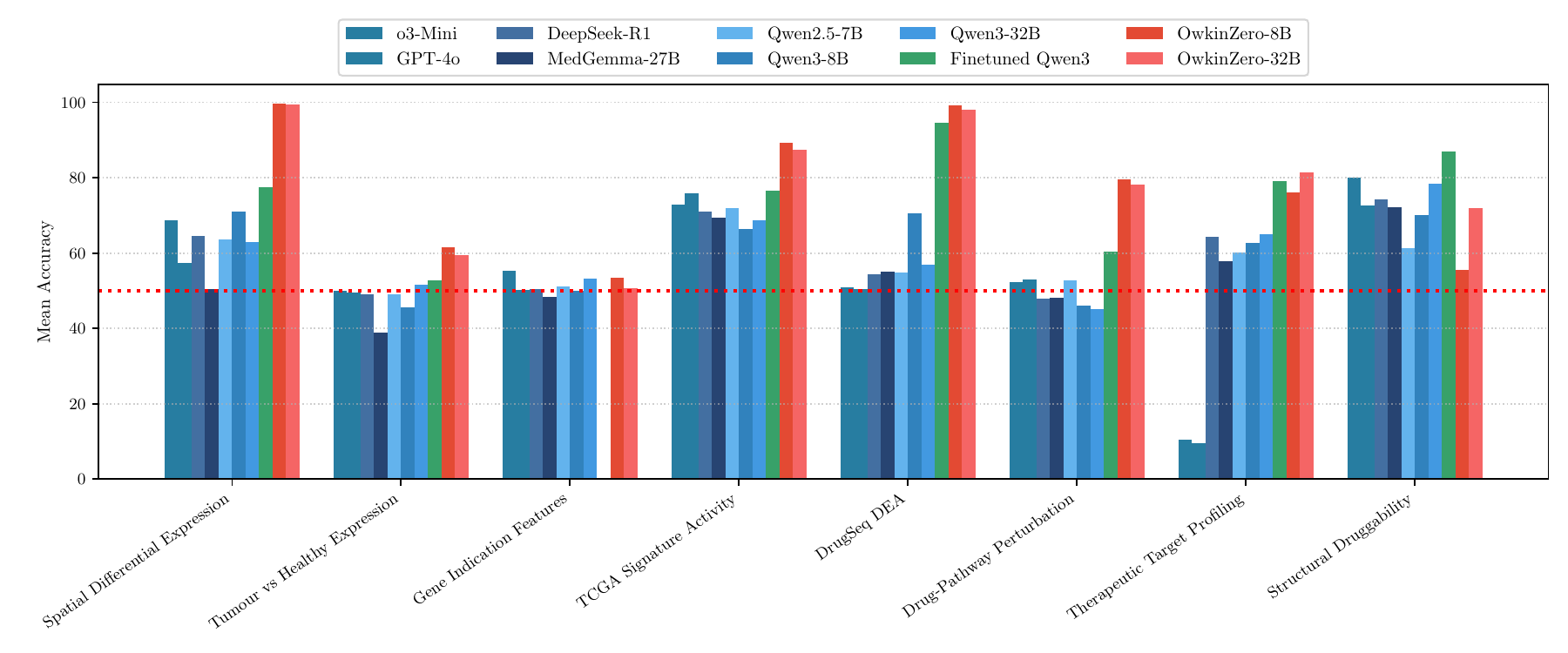}
    \caption{Performance of base, fine-tuned specialist, and OwkinZero models on the biological datasets.}
    \label{fig:barplot}
\end{figure*}

\begin{figure*}[h]
        \centering
        \begin{subfigure}[b]{0.45\textwidth}
           \centering
           \includegraphics[width=\textwidth]{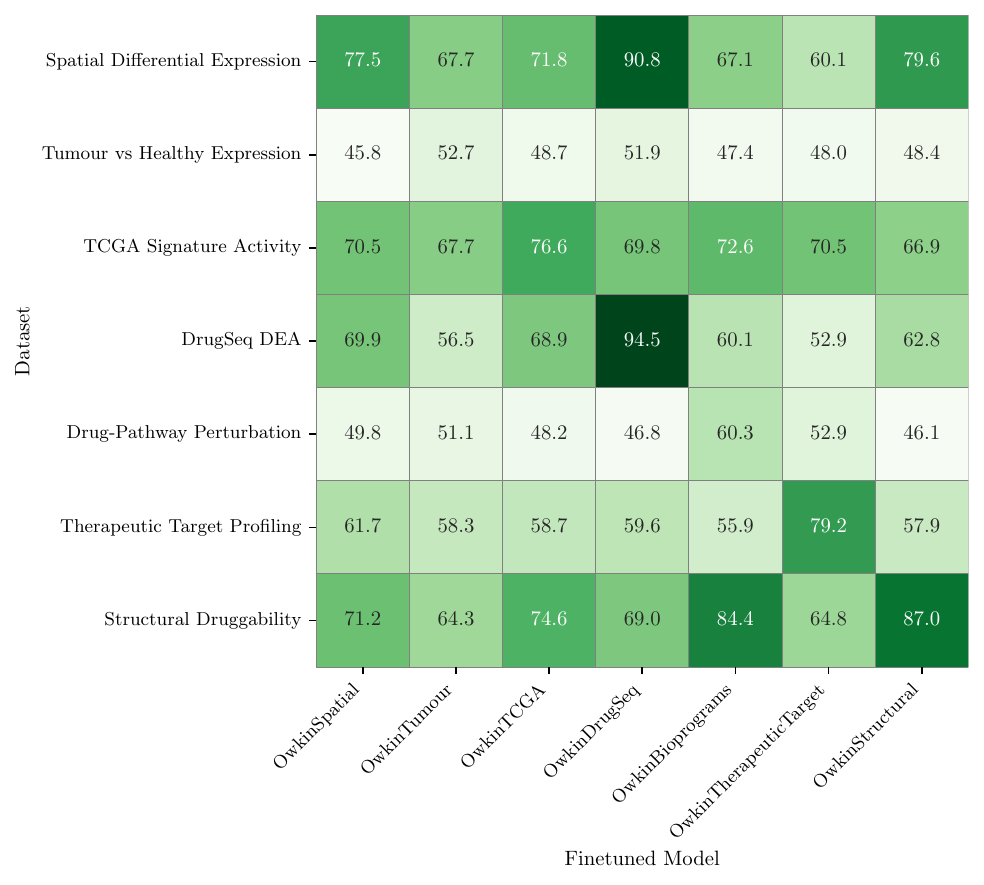}
           \caption{Absolute accuracy In/Out domains}
           \label{fig:confusion-matrix-absolute}
        \end{subfigure}
        \begin{subfigure}[b]{0.45\textwidth}
           \includegraphics[width=\textwidth]{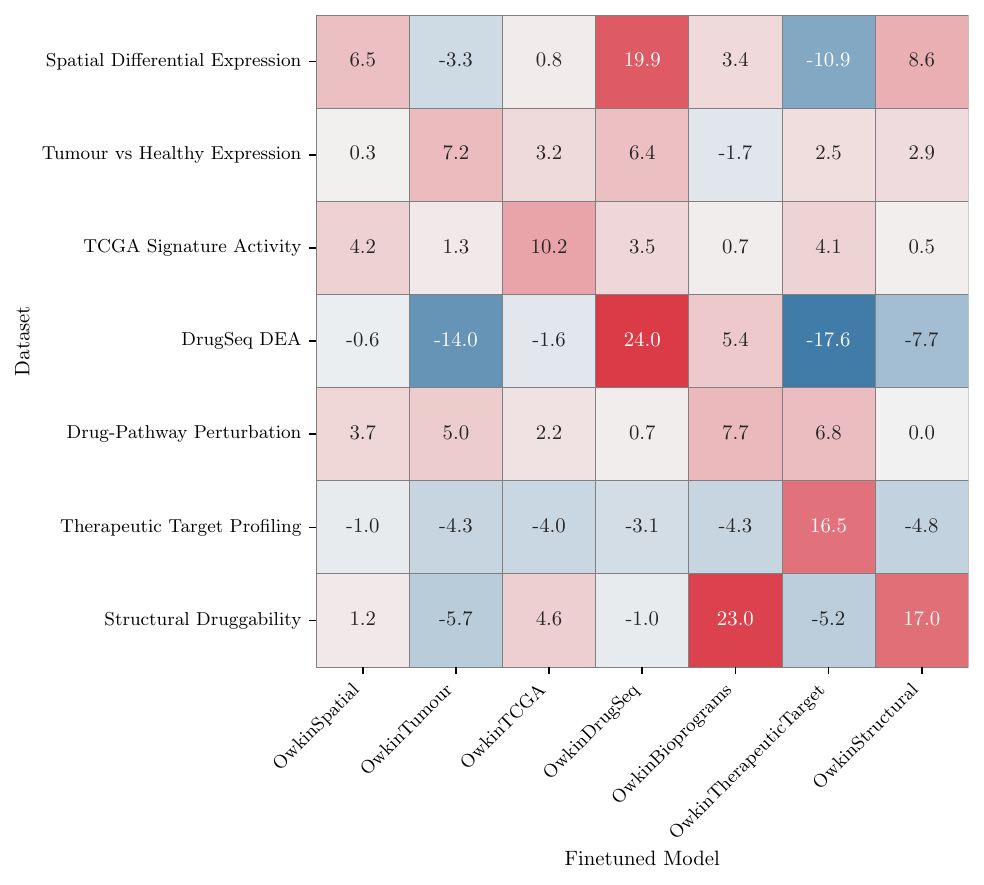}
                 \caption{Relative accuracy In/Out domains}
           \label{fig:confusion-matrix-gain}
        \end{subfigure}
        \caption{Confusion matrices for cross-domain generalisation of specialist models. Left: absolute accuracy. Right: accuracy gain/loss relative to the corresponding base model used for fine-tuning.}
        \label{fig:confusion-matrix}
     \end{figure*}

\begin{figure*}[h]
    \centering
    \begin{subfigure}[b]{0.24\textwidth}
        \centering
        \includegraphics[width=\textwidth]{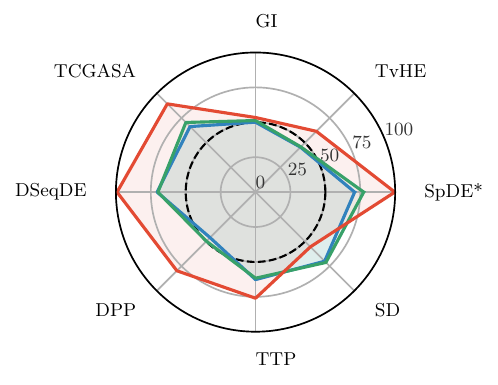}
        \caption{OwkinSpatial}
        \label{fig:spde}
    \end{subfigure}
    \begin{subfigure}[b]{0.24\textwidth}
        \centering
        \includegraphics[width=\textwidth]{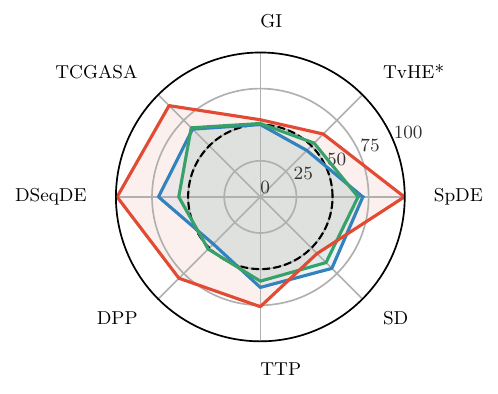}
        \caption{OwkinTumour}
        \label{fig:tvsh}
    \end{subfigure}
    \begin{subfigure}[b]{0.24\textwidth}
        \centering
        \includegraphics[width=\textwidth]{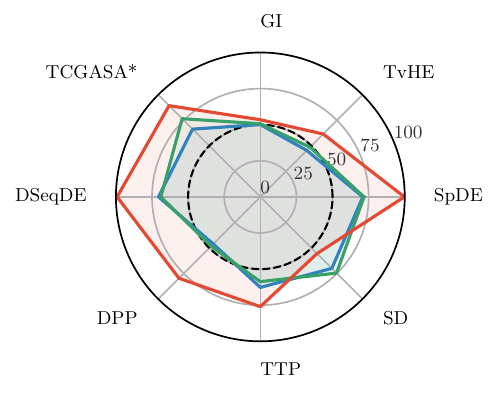}
        \caption{OwkinTCGA}
        \label{fig:tcga}
    \end{subfigure}
    \begin{subfigure}[b]{0.24\textwidth}
        \centering
        \includegraphics[width=\textwidth]{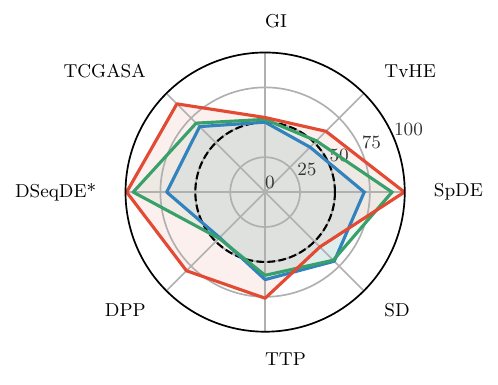}
        \caption{OwkinDrugSeq}
        \label{fig:dseqde}
    \end{subfigure}
    \\
    \begin{subfigure}[b]{0.24\textwidth}
        \centering
        \includegraphics[width=\textwidth]{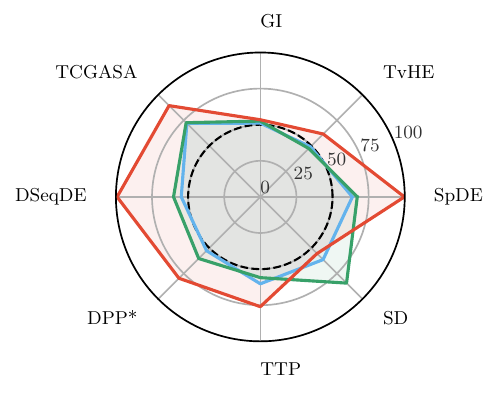}
        \caption{OwkinBioprograms}
        \label{fig:dgppa}
    \end{subfigure}
    \begin{subfigure}[b]{0.24\textwidth}
        \centering
        \includegraphics[width=\textwidth]{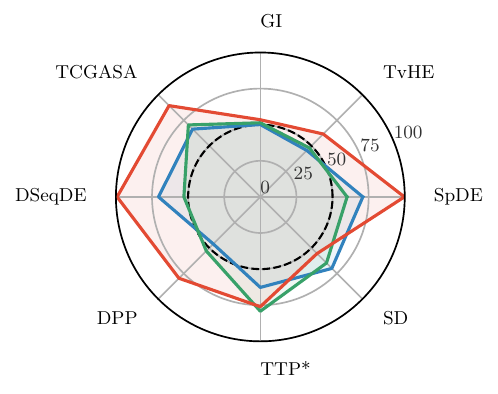}
        \caption{OwkinTherapeuticTarget}
        \label{fig:td}
    \end{subfigure}
    \begin{subfigure}[b]{0.24\textwidth}
        \centering
        \includegraphics[width=\textwidth]{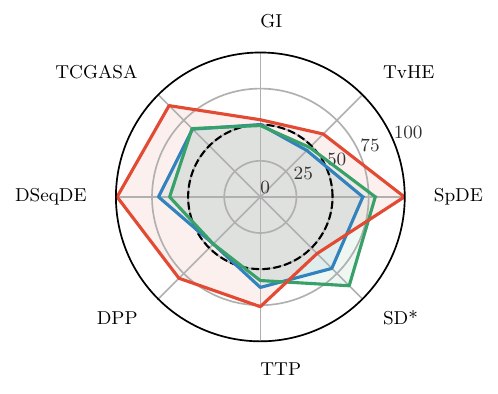}
        \caption{OwkinStructural}
        \label{fig:sd}
    \end{subfigure}
    \begin{subfigure}[b]{0.24\textwidth}
        \centering
        \includegraphics[width=\textwidth]{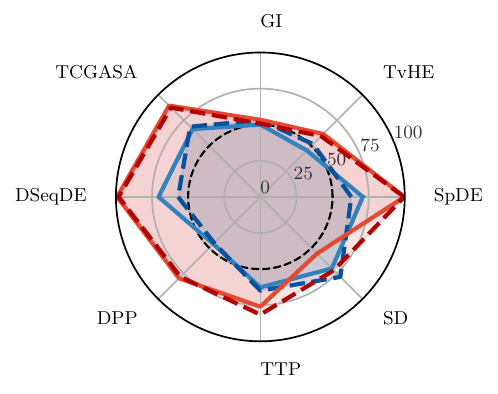}
        \caption{OwkinZero}
        \label{fig:owkinzero}
    \end{subfigure}
    \caption{
        $\raisebox{0.6ex}{\textcolor[HTML]{3182bd}{\rule{1.2em}{1.2pt}}}$~Qwen3-8B
        \hspace{0.5em}
        $\raisebox{0.6ex}{\textcolor[HTML]{38a169}{\rule{1.2em}{1.2pt}}}$~Specialist (fine-tuned)
        \hspace{0.5em}
        $\raisebox{0.6ex}{\textcolor[HTML]{e34a33}{\rule{1.2em}{1.2pt}}}$~OwkinZero 8B
        \hspace{0.5em}
        $\raisebox{0.6ex}{\textcolor[HTML]{08519c}{\rule{0.4em}{1.2pt}\,\rule{0.4em}{1.2pt}\,\rule{0.4em}{1.2pt}}}$~Qwen3-32B
        \hspace{0.5em}
        $\raisebox{0.6ex}{\textcolor[HTML]{b30000}{\rule{0.4em}{1.2pt}\,\rule{0.4em}{1.2pt}\,\rule{0.4em}{1.2pt}}}$~OwkinZero 32B
        \\
        Performance breakdown by dataset for all models. Each subfigure highlights the performance of a specialist 8B model fine-tuned on a single dataset,
        versus OwkinZero 8B model and the base model. The training dataset is starred, except for the subfigure on the bottom right which shows the difference between the 8B and 32B models, for the base models versus the OwkinZero trained on all datasets except \gls{sd}.
    }
    \label{fig:spider-plot}
\end{figure*}

We first benchmarked publicly available models on our datasets. Surprisingly, we noticed that all the models we tried were struggling against our \gls{qa} datasets as shown in~\cref{fig:barplot}. Notably, most models' performances are on par with a random baseline, with the exception of the \gls{sd} dataset, as well as the \glsxtrshort{tcgasa}. More interestingly, no clear hierarchy appears between model sizes, nor between models trained on Math/Code reasoning tasks, which intuitively suggests first that poor performances could be not related to a lack of prior knowledge, and second that reasoning patterns from Math and Code tasks are not directly transferable to biology. %TODO target specific values to back those quite strong claims] 

Next, we investigated whether poor performance is directly related to lack of signal in our \gls{qa} datasets, or if \gls{rl} training could improve performance. We split each dataset into train and test sets, leveraging biological insight to avoid leakage from train to test, and fine-tuned Qwen3-8B models using \gls{grpo} on each dataset. As shown in~\cref{fig:barplot}, the resulting fine-tuned models outperform all baseline models on the test sets, in spite of their relatively small size. 

Finally we fine-tuned 2 models, respectively from Qwen3-8B-Instruct and its 32B version, on all our datasets. We exclude the \gls{sd} from our mixture dataset to allow for investigation on out of domain generalization, which will be detailed in the next paragraph.

OwkinZero models trained on mixture datasets exhibit several interesting behaviors. First of all, in spite of the fact that we performed only one RL training phase (while it is known that further training with supervised finetuning followed by a second round of RL training improves performances substantially), we achieve state-of-the-art accuracy in all in-domain tasks. Moreover, for all the tasks at hand, the training on the mixture dataset allows OwkinZero to reach better performances than the specialist models fine-tuned on their respective tasks, raising hope of cross task generalization. However, for the one dataset held out from our training mixture, \gls{sd}, we observe a common behavior with \gls{rl} training, namely catastrophic forgetting, in the sense that base models perform better than our OwkinZero models on the out-of-domain task.

A finer-grained analysis of the cross-domain generalization ability of the specialist OwkinZero models reveal interesting insights. 
The confusion matrix in~\cref{fig:confusion-matrix-absolute} highlights that, as expected, each specialist model outperforms the other models on the task it was trained on but surprisingly, for some datasets, namely \gls{dseqde} and \gls{dpp}, the respective fine-tuned models also show significant performance improvement over the base model on out-of-domain tasks, specifically \gls{spde} and \gls{sd}. 
Catastrophic forgetting is also present; for instance, the OwkinTumour model's performance drops on the \gls{dseqde} task, while asymmetrically the model trained on \gls{dseqde} shows a small improvement on performance on the \gls{tvhe} task. 

All our results are summarized in~\cref{tab:results}, which highlights that the \gls{gi} task remains unsolved, raising the question of whether there is any signal in the \gls{qa} data, or if the model lacks the prior knowledge in its pre-training to handle such a task. 
\begin{table*}[h]
    \centering
    \begin{tabular}{lcccccccc|c}
        \toprule
        \textbf{Model}&\glsxtrshort{spde}&\glsxtrshort{tvhe}&\glsxtrshort{gi}&\glsxtrshort{tcgasa}&\glsxtrshort{dseqde}&\glsxtrshort{dpp}&\glsxtrshort{ttp}&\glsxtrshort{sd}&All\\
        \midrule
        o3-Mini     &68.6&50.0&\textbf{55.2}&72.93&51.0&52.4&10.5*&\underline{80.0}&55.08 \\
        GPT-4o      &57.4&49.6&50.2&\underline{75.78}&50.5&\underline{52.9}&9.6*&72.6&52.32 \\ 
        DeepSeek-R1 &64.47&49.17&50.42&71.0&54.33&47.92&64.33&74.33&59.5 \\
        MedGemma-27B&50.55&38.83&48.33&69.34&55.0&48.17&57.92&72.17&55.04 \\
        Qwen2.5-7B  &63.74&49.08&51.08&71.94&54.75&52.67&60.17&61.42&58.11 \\
        Qwen3-8B    &\underline{70.97}&45.5&50.08&66.36&\underline{70.5}&46.08&62.67&70.0&\underline{60.27} \\
        Qwen3-32B   &63.0&\underline{51.67}&53.17&68.75&56.92&45.17&\underline{64.92}&78.33&60.24 \\
        \midrule
        OwkinSpatial          &77.47&45.83&51.25&70.54&69.92&49.75&61.67&71.25&62.21 \\
        OwkinTCGA             &71.8 &48.67&50.67&\underline{76.58}&68.9 &48.25&58.67&74.6&62.27 \\
        OwkinTumour           &67.67&\underline{52.67}&50.83&67.67&56.5 &51.08&58.33&64.3&58.63 \\
        OwkinDrugSeq          &\underline{90.84}&51.92&52.08&69.85&\underline{94.5}&46.75&59.58&69.0&\underline{66.81} \\
        OwkinBioprograms      &67.1 &47.42&\underline{52.5}&72.65&60.1 &\underline{60.33}&55.9&84.4&62.55 \\
        OwkinTherapeuticTarget&60.07&48.0&51.33&70.46&52.92&52.92&\underline{79.17}&64.75&59.95 \\
        OwkinStructural       &79.6 &48.42&49.42&66.88&62.8 &46.08&57.9 &\underline{\textbf{87.0}}&62.26 \\
        \midrule
        OwkinZero-8B&\textbf{99.73}&\textbf{61.58}&\underline{53.42}&\textbf{89.36}&\textbf{99.17}&\textbf{79.67}&76.0&55.5&76.8 \\
        OwkinZero-32B&99.54&59.5&50.75&87.43&98.08&78.25&\textbf{81.5}&\underline{71.92}&\textbf{78.37} \\
        \bottomrule
\end{tabular}%
\caption{Results on the biological datasets. Best performances are highlighted in bold. Best performance per categories, namely base models, specialist models, and OwkinZero models, are underlined. If closed source model refuses to anwsers the question for safety reasons, we count it as a failure, which explain certain very low accuracy values for binary tasks.\label{tab:results}}
\end{table*}
\subsection{Reasoning preference and Faithfulness}

\begin{figure*}[h]
    \centering
    \begin{subfigure}[b]{0.3\textwidth}
        \centering
        \includegraphics[width=\textwidth]{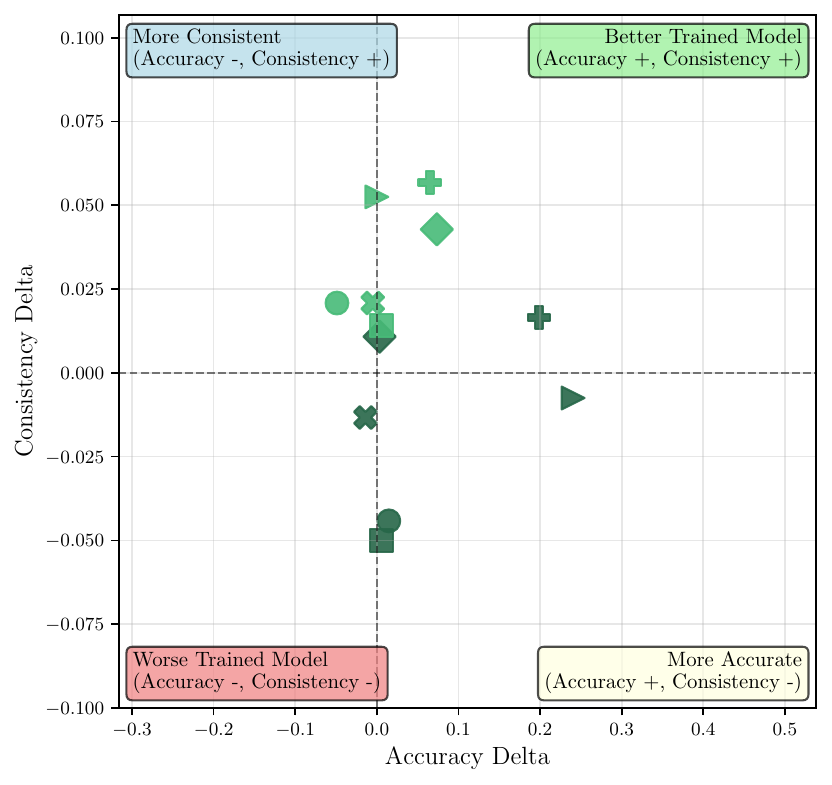}%
        \caption{Accuracy vs Confidence}%
        \label{fig:acc_vs_conf}
    \end{subfigure}
    \begin{subfigure}[b]{0.3\textwidth}
        \centering
        \includegraphics[width=\textwidth]{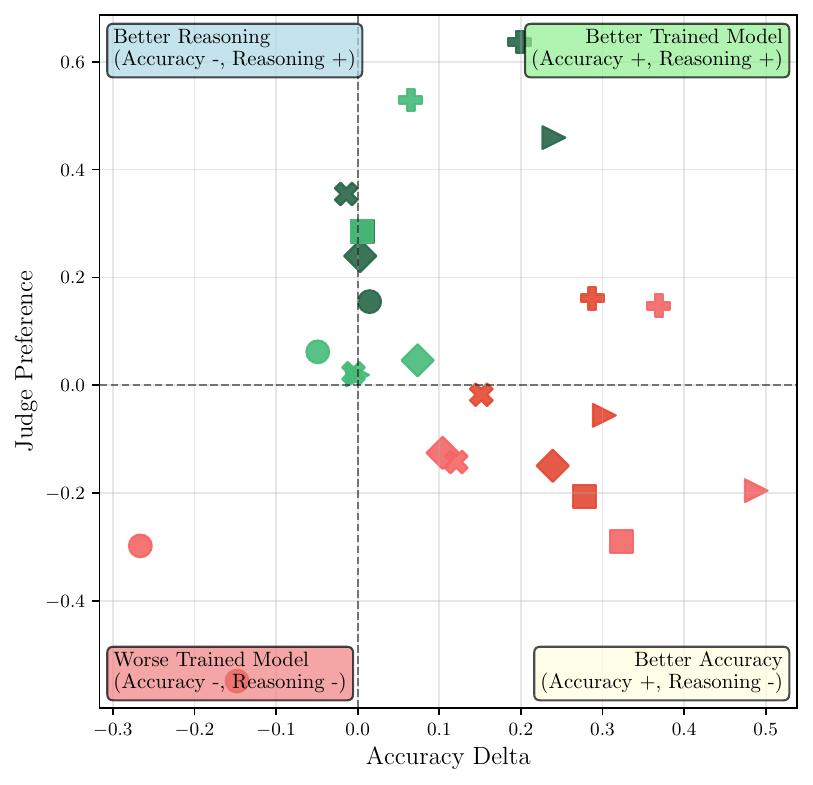}%
        \caption{Accuracy vs Judge preference}%
        \label{fig:acc_vs_judge}
    \end{subfigure}
    \begin{subfigure}[b]{0.3\textwidth}
        \centering
        \includegraphics[width=\textwidth]{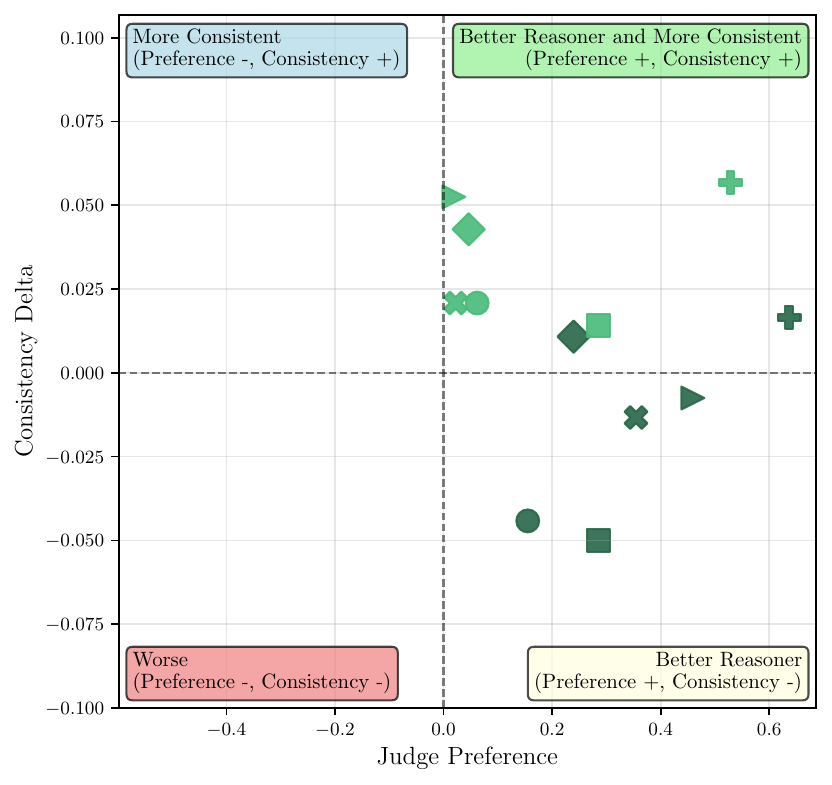}%
        \caption{Judge preference vs Consistency}%
        \label{fig:pref_vs_cons}
    \end{subfigure}
    \caption{
        Each point represents a fine-tuned model evaluated against its corresponding base model. The accuracy and consistency deltas reflect the difference in these metrics between the fine-tuned and base models; positive values indicate improved performance after fine-tuning. For the judge preference, each model is given one point when preferred by the \gls{llm} judge, and the reported metric is averaged over all the completions. Colors denote different models, while marker shapes correspond to the datasets used for evaluation. \\
    \textbf{Models: }
    {\textcolor[HTML]{f56565}{\large$\bullet$}}\,OwkinZero-32B \quad
    {\textcolor[HTML]{e34a33}{\large$\bullet$}}\,OwkinZero-8B \quad
    {\textcolor[HTML]{276749}{\large$\bullet$}}\,OwkinDrugSeq \quad
    {\textcolor[HTML]{48bb78}{\large$\bullet$}}\,OwkinSpatial
    \\
    \textbf{Datasets:}
    $\heavyplus$\glsxtrshort{spde} \quad
    $\blacklozenge$\glsxtrshort{tcgasa} \quad
    $\blacktriangleright$\glsxtrshort{dseqde} \quad
    $\blacksquare$\glsxtrshort{dpp} \quad
    $\heavycross$\glsxtrshort{ttp} \quad
    \large$\bullet$\glsxtrshort{sd}}
    \label{fig:reasoning-quality-preference}
\end{figure*}

Based on those encouraging results, we further investigated whether improvements in model accuracy correlate with enhanced reasoning capabilities. We looked in particular at two dimensions of reasoning: reasoning quality, defined as the standalone coherence of the reasoning and the factuality of the intermediary steps used in reasoning, and reasoning faithfulness, understood as the coherence between the model’s reasoning and its final answer. 

For both of these measures, we relied on baseline comparative metrics using \gls{llm} as a judge. We picked SOTA models in the 32B size range, and used them to evaluate our reasoning traces. Faced with the intricacies of the issue of reasoning evaluation for non-formal reasoning traces (i.e., different from math and code), we resolved to focus on simplicity. While the chosen methods come with limitations, as detailed later, they represent a first simple and flexible way of measuring reasoning quality. 

For reasoning quality, we focused on a proxy of quality - preference. We approximate that better quality reasoning will be preferred over lower quality reasoning by a competent \gls{llm} judge. We presented the Judge \gls{llm} with a question and two alternative thinking traces, without providing the actual answer in the prompt. We then requested a binary preference score (see~\cref{supplementary:judge-prompt} for the detailed prompt used). To ensure the robustness of the evaluation, each judgment was performed five times.

We furthermore examine the consistency between the reasoning trace generated by the model and its final chosen answer. It is indeed an emerging fact in the reasoning literature, that reasoning models, despite their better accuracy, do not output reasoning traces in accordance with their final answer~\citep{paul2024makingreasoningmattermeasuring, turpin_language_2023}. Examples showed models reaching the correct answer but with a contradictory or illogical thought process.

\begin{quote}
\begin{lstlisting}[basicstyle=\small\ttfamily,breaklines=true]
"[...]Therefore, based on the current knowledge, there's no direct evidence linking CHEK1 inhibition to AURKA gene deregulation. So the answer would be B, No.
</think>
<answer>
A
</answer>"
\end{lstlisting}
\end{quote}

To assess this, a Judge LLM was also employed to evaluate whether the reasoning provided was consistent with the selected answer. This was done by providing the Judge \gls{llm} with only the reasoning trace, and asking what the most likely answer was based on this reasoning alone. We then compared this answer with the provided answer, to see if they matched. 

The preference results, shown in~\cref{fig:acc_vs_judge}, show two emerging results. First, reasoning quality seems to be linked to better accuracy for models fine-tuned on a single dataset. Second, mixture model’s increase in accuracy does not seem to be reflected in reasoning quality. Given the limitations of our approach, we consider this evidence anecdotal at this stage and plan to further investigate these emerging facts. 

The faithfulness analysis showed~\cref{fig:acc_vs_conf} that models fine-tuned with single in-domain datasets maintained high consistency between reasoning and answers while improving accuracy on those specific datasets. Conversely, models trained on a mixture of datasets, despite achieving higher overall accuracy, often exhibited significantly worse consistency between their reasoning and their answers. This aligns with findings from other research~\citep{deepseek-ai_deepseek-r1_2025} suggesting that initial \gls{rl} passes can improve accuracy but generate low quality reasoning traces, necessitating further \gls{sft} and subsequent \gls{rl} steps for alignment.

Finally, we observe a positive correlation between reasoning quality and reasoning consistency, as shown in~\cref{fig:acc_vs_judge}.
This evidence can only be considered anecdotal at this stage, due to the inherent limitation of \gls{llm} judges. The reliability of these AI-based evaluators in accurately assessing the nuances of sophisticated reasoning remains a significant concern. In particular, one can ask if a model not able to accurately answer a question can judge its answer. Manual inspection of the preference data further tends to highlight that more assertive answers (no ``maybe'', ``perhaps'', \ldots) seem to have judge preferences, alluding to the superficial nature of the judging criteria, despite our careful prompting. 

\section{TRAINING FRAMEWORK}%
\label{sec:training-framework}

In this section, we summarize the training methodology used to fine-tune our models.
We use an updated version of the \gls{grpo} algorithm~\citep{shao_deepseekmath_2024} which was designed to optimize for accuracy on verifiable questions without the need of a critic model, by averaging rewards from multiple generations of the same prompt for the advantage calculation.

A well known issue with the original \gls{grpo} loss is the length bias, where per-response normalization can disproportionately favor shorter sequences and discourage the model from generating longer and more complex chains of thought~\citep{liu_understanding_2025, yu_dapo_2025, mistral-ai_magistral_2025}.
Shorter responses go against recent findings according to which longer completions are essential for models' \textit{reasoning}, allowing more test-time compute according to the question's difficulty~\citep{snell2024scalingllmtesttimecompute, wei2023chainofthoughtpromptingelicitsreasoning}.
We utilized the Hugging Face TRL library's implementation~\citep{von_werra_trl_nodate}, which employs a per-batch token-level normalization strategy.
The modified objective called BNPO normalizes the loss by the total number of tokens across all responses from all prompts within a given batch. 
This ensures that each token contributes equally to the gradient update regardless of the length of the response it belongs to.

The BNPO objective, which we optimize, can be expressed as:

\begin{strip}
\begin{equation*}
\label{eq:bnpo_loss}
\begin{split}
\mathcal{J}_{\text{BNPO}}(\pi_{\theta}) & = \mathbb{E}_{\{\mathbf{q}\}_{n=1}^{N} \sim p_Q, \{o_i^n\}_{i=1}^{G} \sim \pi_{\text{ref}}(\cdot|\mathbf{q}_n)} \\
& \frac{1}{\sum_{n=1}^{N} \sum_{i=1}^{G} |o_i^n|} \sum_{n=1}^{N} \sum_{i=1}^{G} \sum_{t=1}^{|o_i^n|} \left\{ \min \left( r_{n,i,t}(\theta) \hat{A}_{n,i}, \text{clip}(r_{n,i,t}(\theta), 1-\epsilon, 1+\epsilon) \hat{A}_{n,i} \right) \right\}
\end{split}
\end{equation*}
\end{strip}

\noindent where $N$ is the number of prompts in the batch and $G$ is the number of responses per prompt.
The importance sampling ratio $r_{n,i,t}(\theta)$ is defined as:
\begin{equation}
\label{eq:importance_ratio}
r_{n,i,t}(\theta) = \frac{\pi_{\theta}(o_{i,t}^n | \mathbf{q}_n, o_{i,<t}^n)}{\pi_{\text{ref}}(o_{i,t}^n | \mathbf{q}_n, o_{i,<t}^n)}
\end{equation}

\noindent where $o_{i,t}^n | \mathbf{q}_n, o_{i,<t}^n$ represents the generation of the model for the $t$-th token of the $i$-th response for the $n$-th prompt, $\pi_{\theta}$ is the current policy and $\pi_{\text{ref}}$ is the reference policy.

The advantage estimate $\hat{A}_{n,i}$, which is constant for all tokens in a given response, is the group-normalized reward.
For a set of rewards $\{R(o_j^n)\}_{j=1}^{G}\}$ computed for each response in the group for prompt $\mathbf{q}_n$, the advantage is:
\begin{equation}
\label{eq:advantage}
\hat{A}_{n,i} = \frac{R(o_i^n) - \mu_n}{\sigma_n + \epsilon_{\text{std}}}
\end{equation}
where $\mu_n$ and $\sigma_n$ are the mean and standard deviation of the rewards within the group for prompt $\mathbf{q}_n$, and $\epsilon_{\text{std}}$ is a small constant to ensure numerical stability.

Note that when the batch size is the same as the number of generations ($N==G$), the BNPO objective becomes equivalent to the DAPO loss~\citep{yu_dapo_2025} where answers share the same question-level normalizer.

Note also that the KL-divergence penalty term, often present in PPO and \gls{grpo} formulations to constrain the policy shift, is deliberately omitted by setting its coefficient $\beta$ to 0, the default value in TRL.
This choice is also supported by recent findings in works such as Magistral~\citep{mistral-ai_magistral_2025} and Open-Reasoner-Zero~\citep{hu_open-reasoner-zero_2025}, which demonstrate that \gls{grpo} is stable enough so that removing the KL regularization achieves on par or better performance, while reducing memory usage and improving training speed due to not having to have the reference model in memory.

\subsection{Implementation Details}

Our training implementation was built on the Hugging Face Open-r1 library~\citep{hugging_face_open_2025} using the \texttt{GRPOTrainer} from TRL~\citep{von_werra_trl_nodate}.
We used a group size of $G=10$ and a batch size of $N=10$.
In line with best practices and to mitigate the risk of catastrophic forgetting, all models were trained for a single epoch over their respective training datasets.
This single-pass approach ensures the model adapts its pre-trained knowledge to the new tasks without destructively overwriting its core capabilities.
All other hyperparameters were kept to the default settings of the \texttt{GRPOTrainer}, including setting the $\beta$ parameter to 0 to disable the KL penalty.

We used the following reward functions with equal weights, which are partially inspired by the open-r1 repository:
\begin{itemize}
    \item \texttt{Format}: 1 if the reasoning process is enclosed within <think> and </think> tags, while the final answer is enclosed within <answer> and </answer> tags, 0 otherwise.
    \item \texttt{Tag Count}: A sum of 0.25 for each of <think> and </think>, <answer> and </answer> tags present exactly once in the completion, summing to 1 with correct formatting.
    \item \texttt{Multiple choice (valid choice)}: 1 if the answer string exactly matches one of the possible choices, 0 otherwise.
    \item \texttt{Multiple choice (correct answer)}: 1 if the answer string exactly matches the correct answer, 0 otherwise.
\end{itemize}

We trained all models on 16 H200 GPUs over the course of 2 months.
The mixture of datasets created for the OwkinZero models was created by sampling around 5,000 samples for each question type.
Each model required less than 24 hours to train, the longest being the 32B model trained on the mixture of datasets which took around 18 hours on 8 H200 GPUs.

\section{RELATED WORK}

Our work sits alongside a growing body of research applying advanced AI to biomedical sciences.

The most popular recent efforts have focused on developing agentic systems that can query the data and build answers from various evidence like Biomni~\citep{huang_biomni_2025}, but without improving the models' intrinsic biological reasoning.
Another way to improve an \gls{llm}'s ability to answer biological questions without having to modify its weights is to use a fixed \gls{llm} component and a trainable encoder, like ChatNT~\citep{de_almeida_multimodal_2025} for biological sequences, which is trained on a set of biological tasks derived from a foundation model benchmark~\citep{dalla-torre_nucleotide_2025}.

Recent fine-tuned \glspl{llm} for biomedical tasks include Google Deepmind's TxGemma~\citep{wang_txgemma_2025} and MedGemma~\citep{sellergren_medgemma_2025}.
TxGemma is a suite of open-weights fine-tuned models for various biomedical tasks based on the the Therapeutics Data Commons~\citep{huang_therapeutics_2021}.
However, it is trained with supervised fine-tuning, greatly limiting the ability of the models and requiring the release of separate `predictive models' and `conversational models'.
MedGemma is a collection of medical vision-language foundation models, and it is also mainly trained with supervised fine-tuning, \gls{rl} being mainly used for the multimodal aspect.

In contrast, \gls{rlvr} is starting to show promise as a powerful tool to improve \glspl{llm}'s accuracy on specific tasks like gene classification~\citep{swanson_rl-finetuning_2025}, various chemistry problems~\citep{narayanan_training_2025}, and protein design~\citep{hla_pro-1_2025}.
Furthermore, Magistral by MistralAI~\citep{mistral-ai_magistral_2025} demonstrated that even for smaller base models, \gls{rl} alone can develop similar or better performance than distillation of a much larger model, paving the way for its applications to various domains.

\section{DISCUSSION}%
\label{sec:discussion}

In this paper, we demonstrate for the first time that language models can be adapted to perform complex biological problem-solving, with important potential applications in biomedical research, in particular therapeutics discovery. We build a new kind of benchmark for biology reasoning by derived \gls{qa} pairs covering several scales of biology and critical components of the drug discovery process.

Our work demonstrates that moderately-sized language models (8--32B parameters), when fine-tuned with a single phase of reinforcement learning on a curated suite of verifiable biological tasks, can consistently outperform larger, general-purpose commercial systems.
This finding provides strong evidence that for specialized scientific domains, expert data curation and targeted alignment can be more impactful than model scale alone, a conclusion supported by similar findings in adjacent domains~\citep{swanson_rl-finetuning_2025, narayanan_training_2025}.

Our results confirm that \gls{rl} significantly boosts in-domain accuracy across a range of datasets.
However, the degree of improvement was not uniform.
Tasks such as the \gls{gi} dataset, which requires reasoning over complex features like intra-tumour variability, proved to be too hard for our training framework.
This heterogeneity in task difficulty is not surprising, it is also reported in other broad-scope benchmarks and suggests that some reasoning skills may be particularly reliant on specific knowledge absent from pre-training or may exceed the reasoning capacity of the base models~\citep{narayanan_training_2025, wang_txgemma_2025}.
This variability extends to our analysis of generalization which reveals a more complex picture: while out-of-domain generalization was observed on some tasks, most fine-tuned models were prone to catastrophic forgetting.
This suggests that "biological reasoning" is not a monolithic capability but a combination of diverse skills, some of which may require more targeted training strategies to develop and retain.

A key limitation of our RL-only approach, which is optimized primarily for accuracy, is that it is insufficient on its own to guarantee faithful reasoning~\citep{paul2024makingreasoningmattermeasuring, turpin_language_2023}.
This was particularly evident in our mixture-trained models, which improved overall accuracy at the cost of reasoning faithfulness.
This trade-off motivates a clear path for future work, moving beyond a single \gls{rl} phase towards multi-stage recipes that include supervised fine-tuning on curated chains of thought.
The use of multiple-choice questions is also a limiting factor, open-ended questions are likely to be more challenging for the model and teach it to reason about the question and answer in a more free-form way~\citep{narayanan_training_2025}. 

We developed a new kind of benchmark on which one can evaluate language models’ ability to answer complex biological questions. In future work, we will scale the size and diversity of our benchmark, by developing robust alternatives to LLM-as-a-judge. We believe that such benchmarks are essential for tackling the next frontier in AI-driven biology discovery. 

This study represents a first step towards developing a fundamentally new kind of biological reasoning capability that we believe is essential for enabling breakthrough discoveries in drug development. The experimental landscape in biology is inherently sparse and fragmented: perturbation assays, for example, are typically constrained to specific panels of in vitro cell lines and limited gene sets—contexts that fail to capture the full complexity of human biology. Consequently, a critical bottleneck in the field lies in the ability to bridge these experimental gaps and extrapolate insights across sparse and disparate biological contexts~\citep{lotfollahi_scgen_2019, wenkel_txpert_2025}.

We note that advanced agentic systems like Biomni~\citep{huang_biomni_2025}, if equipped with the right tools and access to the right data, is an orthogonal way to achieving high performances on our novel benchmark. Indeed, such systems can iteratively load, analyze and process data to refine their outputs. In contrast, our approach  focuses on enhancing the model's intrinsic reasoning capabilities to generalize beyond the limits of current LLMs relying on prior knowledge solely. This reasoning-centered step precludes genuine scientific discovery and mirrors the scientific process in biomedicine, where researchers begin with intuition grounded in expertise before proceeding to wet-lab experimentation to observe data.

The most transformative AI systems for drug discovery will ultimately require the combination of both paradigms: advanced reasoning models that can select and run appropriate tools, have access to rich datasets, and are also able to reason about the results of such tools to answer complex questions and propose novel experiments~\citep{jin_search-r1_2025, qian_toolrl_2025}.

\subsubsection*{Acknowledgements}
\label{sec:acknowledgements}
We are grateful to the participants of the Owkin internal hackathon for their valuable contributions to the design and curation of datasets presented in this work.
In particular we thank Barbara Bodinier and Khalil Ouardini for inspiring the \gls{dseqde} dataset, Alexandre Grimaldi, Roberta Codato and Caroline Hoffmann for the \gls{tvhe} and \gls{gi} datasets, Christian Esposito, Gaëtan Dissez, Maxime Touzot and Alice Mac Kain for the \gls{ttp} dataset, Thomas Mathieu, Xenia Snetkov, Yacine Bareche and Almudena Espin Perez for the \gls{spde} dataset, and Anna Song, Gergana Bounova and Antoine Simon for the \gls{sd} dataset.
We also thank Elodie Pronier and Jean-Philippe Vert for general guidance.

We would also like to thank Quentin Gallouédec for a valuable discussion as well as all HuggingFace contributors for their high-quality open source implementation of TRL, Michael Hla for an insightful conversation on applying \gls{rl} to biological reasoning, and Yann Fleureau for his expert advice on scaling \gls{rl} for improved reasoning.
The authors would like to thank Ginkgo Bioworks for providing the proprietary drug perturbation assays.

This study also makes use of data generated by the MOSAIC consortium (Owkin; Charité – Universitätsmedizin Berlin (DE); Lausanne University Hospital - CHUV (CH); Universitätsklinikum Erlangen (DE); Institut Gustave Roussy  (FR); University of Pittsburgh (USA)).

\bibliographystyle{plainnat}
\bibliography{owkinzero.bib}

%%%%%%%%%%%%%%%%%%%%%%%%%%%%%%%%%%%%%%%%%%%%%%%%%%%%%%%%%%%%

\newpage
\onecolumn

\aistatstitle{OwkinZero \\
Supplementary Materials}

\setcounter{section}{0}
\renewcommand{\thesection}{S\arabic{section}}
\makeatletter
\renewcommand{\theHsection}{S\arabic{section}}
\makeatother

\section{Dataset Details}\label{sec:suppdatasets}

In this section, we provide detailed descriptions of each dataset used in our study, including the curation pipeline, question types, schema, representative examples, and the strategy used for constructing train/test splits.

\subsection{\glsfirst{spde}}
\paragraph{Description}
The \glsxtrshort{spde} dataset probes spatial transcriptomic heterogeneity between \emph{tumour islets} and \emph{stroma} across multiple cancer indications using MOSAIC Visium data~\citep{mosaic_consortium_mosaic_2025}. Each item is a two-option multiple-choice Q\&A that asks which gene is \emph{upregulated} or \emph{downregulated} in tumour islets relative to the stromal compartment for a given indication. 

\paragraph{Curation Strategy}
From MOSAIC features we extracted the spatial DEA score for the tumour-islet vs stroma contrast. We discarded entries with missing scores and restricted questions to two forms: \emph{Upregulated in tumour islets versus stroma} and \emph{Downregulated in tumour islets versus stroma}.

For each indication, we sampled extreme examples from the tails of the score distribution and paired them with non-extreme distractors from the same indication. Concretely, given scores \(s\) and letting \(Q_p\) denote the \(p\)-quantile:
\begin{itemize}
\item \textbf{Downregulated:} pick genes with $s \le Q_{0.01}(s)$ as extremes; choose distractors with $s \ge -0.5$.
\item \textbf{Upregulated:} pick genes with $s \ge Q_{0.99}(s)$ as extremes; choose distractors with $s \le 0.5$.
\end{itemize}
We used sampling with replacement to reach a target number of pairs per indication. Answer letter (A/B) assignment was randomized per item. All items include the indication in the question stem and present options as \texttt{HGNC (ensembl ENSG...)} for clarity and disambiguation.

\paragraph{Q\&A Types}
Two binary selection tasks:
\begin{enumerate}
    \item Upregulated in tumour islets versus stroma
    \item Downregulated in tumour islets versus stroma
\end{enumerate}

\paragraph{Q\&A Schema}
\begin{quote}
\ttfamily
Which gene is \{upregulated/downregulated\} in tumour islets versus stroma in \{INDICATION\}?\\
A) \{HGNC\_A\} (ensembl \{ENSG\_A\}) \quad B) \{HGNC\_B\} (ensembl \{ENSG\_B\})\\
\textbf{Answer:} \{A/B\}
\end{quote}

\paragraph{Q\&A Example}
\begin{quote}
\textbf{Q:} Which gene is downregulated in tumour islets versus stroma in Bladder urothelial carcinoma?\\
A) AFAP1 (ensembl ENSG00000196526)\quad B) BPIFA1 (ensembl ENSG00000198183)\\
\textbf{A:} B
\end{quote}

\paragraph{Train/Test Split Strategy}
We constructed a conservative split to prevent leakage of decision cues:
\begin{itemize}
    \item No \emph{indication} or \emph{gene} shared across splits. The train set is composed of the following indications : ``Bladder urothelial carcinoma'', ``Lung adenocarcinoma'', ``Ovarian serous cystadenocarcinoma'', ``Bladder urothelial carcinoma'' and ``Mesothelioma''. The test set is only composed of ``Invasive breast carcinoma'', and no genes are shared between the two sets.
    \item Randomized A/B position independent across splits to avoid answer-letter shortcuts.
\end{itemize}

% ---------------------------

\subsection{\glsfirst{tvhe}}
\paragraph{Description}
This dataset targets indication-level transcriptomic differences between \emph{tumour} and \emph{adjacent normal} tissues across TCGA cohorts. Each item is a two-option multiple-choice Q\&A asking which tissue type exhibits higher expression for a specific gene in a given cancer type. The questions are deliberately phrased in formal biomedical language to promote domain-specific reasoning rather than surface cue matching. 

\paragraph{Curation Strategy}
Ground truth supervision was derived from matched tumour and adjacent-normal bulk RNA-seq profiles from TCGA. Cancer indications were stratified by retaining only those with at least two adjacent-normal samples. For each selected indication, differential expression analysis was performed using the \texttt{rank\_genes\_groups} function in the \texttt{scanpy} library~\citep{scanpy2018}, applying a non-parametric Wilcoxon rank-sum test to compare gene expression between tumour and normal samples. The function returns multiple-testing corrected $p$-values (Benjamini--Hochberg FDR) and $\log_2$ fold changes (tumour vs.\ normal) as part of its output. Genes were assigned to mutually exclusive sets using the following thresholds:

\begin{itemize}
  \item \textbf{Tumour-up:} $\mathrm{FDR} < 0.05$ and $\log_2\mathrm{FC} > 1$;
  \item \textbf{Normal-up:} $\mathrm{FDR} < 0.05$ and $\log_2\mathrm{FC} < -1$.
\end{itemize}
Genes not meeting these criteria were excluded from question generation. 

In order to generate the Q\&A pairs, first, for every $(\text{indication}, \text{gene})$ in either set, we generated one question:
\begin{quote}
``Is \textbf{GENE} more expressed in \textbf{INDICATION} tumour tissue or in \textbf{INDICATION} adjacent normal tissue?''
\end{quote}

For the answers, options were fixed as A) tumour tissue and B) adjacent normal tissue. The answer key was assigned by set membership (tumour-up $\rightarrow$ A; normal-up $\rightarrow$ B).

Next, to improve linguistic diversity while preserving semantics, we applied controlled rewording with an \gls{llm} helper which did not have access to the questions' answers. The phrasing variations include:
\begin{itemize}
    \item Substituting technical synonyms for core terms: ``transcript abundance'', ``mRNA levels'', ``transcriptional abundance'', ``expression levels''.
    \item Alternating tissue terminology: ``tumour tissue'' / ``neoplastic tissue'' / ``neoplastic cells'' vs.\ ``adjacent normal'' / ``non-neoplastic tissue''.
    \item Changing the grammatical form of the question: from direct yes/no (``Is~GENE~more expressed in...?'') to descriptive (``Does~GENE~exhibit higher...?'') or selection-based (``Which tissue type exhibits higher...?'').
    \item Flipping the tissue order in the question stem while preserving the original answer option order.
\end{itemize}
These linguistic variations require models to extract the comparative logic from the question rather than memorising fixed templates.

\paragraph{Q\&A Types}
One binary comparison task:
\begin{itemize}
    \item \texttt{expression\_tumour\_vs\_healthy}: which tissue (tumour vs.\ adjacent normal) shows higher expression for a given gene and indication?
\end{itemize}

\paragraph{Q\&A Schema}
\textit{(Schema examples; not exhaustive)}
\begin{quote}
\ttfamily
Does \{GENE\} exhibit higher \{transcript abundance / mRNA levels / transcriptional abundance / expression levels\} in \{INDICATION\} \{neoplastic tissue / tumour tissue / neoplastic cells\} compared to the corresponding \{non-neoplastic tissue / adjacent normal tissue\}?\\
A) non-neoplastic tissue \quad B) neoplastic tissue (option ordering shuffled)\\
\textbf{Answer:} \{A/B\}
\end{quote}
\begin{quote}
\ttfamily
In \{INDICATION\}, does \{GENE\} exhibit higher \{transcript abundance / mRNA levels / transcriptional abundance / expression levels\} in the \{neoplastic tissue / tumour tissue / neoplastic cells\} compared to the corresponding \{non-neoplastic tissue / adjacent normal tissue\}?\\
A) tumour tissue \quad B) adjacent normal tissue (option ordering shuffled)\\
\textbf{Answer:} \{A/B\}
\end{quote}

\paragraph{Q\&A Examples}
\begin{quote}
\textbf{Q:} Does HERC3 exhibit higher transcript abundance in papillary renal cell carcinoma (KIRP) neoplastic tissue compared to matched non-neoplastic tissue?\\
A) non-neoplastic tissue B) neoplastic tissue\\
\textbf{A:} A\\

\textbf{Q:} In the context of Lung adenocarcinoma, which tissue type exhibits a higher level of KIAA1328 mRNA abundance?\\
A) Lung adenocarcinoma adjacent normal tissue B) Lung adenocarcinoma tumor tissue\\
\textbf{A:} A\\
\end{quote}

\paragraph{Train/Test Split Strategy}
Like in the \gls{spde} dataset, no indication or gene are shared across splits:
\begin{itemize}
    \item The train set is composed of the followind TCGA indications (for acronym definition, see \url{https://gdc.cancer.gov/resources-tcga-users/tcga-code-tables/tcga-study-abbreviations}) : ``KICH'', ``BRCA'', ``LUAD'', ``THCA'', ``PRAD'', ``COAD'', ``LUSC'', ``KIRP'', ``STAD'', ``HNSC'', ``LIHC'', ``CHOL'', ``GBM'', ``ESCA'', ``BLCA''. The test set is composed of ``KIRC'', ``UCEC'', ``READ''.
    \item The train and test sets share no genes in common.
    \item Stratified by direction (neoplastic\,$>$\,non-neoplastic vs.\ non-neoplastic\,$>$\,neoplastic) to maintain balance.
    \item Rephrasing variants are distributed across both splits to prevent lexical shortcuts.
\end{itemize}

% ---------------------------

\subsection{\glsfirst{gi}}
\paragraph{Description}
The GI dataset contains 127,069 training and 22,484 test True/False questions, each linked to a \emph{(gene, indication)} pair. These questions are derived from a variety of biological feature types spanning:
\begin{itemize}
    \item \textbf{Indication-level expression contrasts} between tumour tissue and diverse reference tissues (e.g., spleen, bone marrow, blood, PBMC, liver).
    \item \textbf{Genomic alteration burden} statements such as the frequency of copy-number variations (CNVs) for a given gene within an indication.
    \item \textbf{Intra-tumour expression variability} across malignant subpopulations (``heterogeneity'' vs.\ ``minimal variability'').
    \item \textbf{Malignant pseudobulk enrichment} (e.g., whether malignant cell pseudobulks \emph{frequently} show elevated expression of a gene).
\end{itemize}
All questions are phrased in formal biomedical terminology (e.g., ``transcript abundance'', ``significantly elevated'', ``frequency of copy number variations'', ``minimal expression variability'') to promote domain-aware reasoning beyond template memorisation. Source data include MOSAIC Bladder visium/single-cell datasets.

\paragraph{Curation Strategy}
\gls{gi} supervision was derived from in-house precomputed feature scores generated from MOSAIC data. These features span multiple biological axes, including gene expression levels, spatial expression patterns, cell-type specificity, and functional signatures such as cathepsin or endocytosis activity. Based on these precomputed scores, a variety of binary question templates were instantiated, reflecting structured mappings from feature space to question space. Below is a representative subset of the question types derived from these features:

\begin{itemize}
  \item Does \texttt{[gene]} in \texttt{[indication]} have high expression in malignant cells?
  \item Does \texttt{[gene]} in \texttt{[indication]} show high expression in tumours enriched for the cathepsin signature?
  \item Does \texttt{[gene]} in \texttt{[indication]} display spatial autocorrelation of expression?
  \item Does \texttt{[gene]} in \texttt{[indication]} show elevated expression in tumour core relative to tumour edge?
  \item Does \texttt{[gene]} in \texttt{[indication]} exhibit higher expression in tumour versus adjacent normal tissue?
  \item Does \texttt{[gene]} in \texttt{[indication]} have a higher proportion of expression in malignant compared to stromal cells?
\end{itemize}

Following template instantiation, similar to \gls{tvhe} dataset curation, a controlled rephrasing procedure was applied to increase linguistic diversity. This involved generating multiple semantically equivalent variants of each question. For example, a question of the form:

\begin{quote}
\ttfamily
Does \texttt{[gene]} in \texttt{[indication]} have a higher expression in tumour versus spleen?
\end{quote}

was rephrased as:

\begin{quote}
\ttfamily
Is the gene expression level of \texttt{[gene]} significantly elevated in \texttt{[indication]} tumour tissue compared to normal spleen tissue?
\end{quote}

Similarly, questions about genomic alterations such as:

\begin{quote}
\ttfamily
Does \texttt{[gene]} in \texttt{[indication]} have a high proportion of patients with copy number alterations for this gene?
\end{quote}

were reworded to:

\begin{quote}
\ttfamily
Is the frequency of copy number variations in the \texttt{[gene]} gene significantly elevated in \texttt{[indication]} patient samples?
\end{quote}

Like in the \gls{tvhe} dataset, this rephrasing was done with an \gls{llm} which did not have access to the questions' answers.

\paragraph{Q\&A Types}
Table~\ref{tab:gi_qtypes} enumerates the full set of question types derived from the feature scores.

\begin{longtable}{@{}p{0.48\textwidth}@{}p{0.48\textwidth}@{}}
\toprule
\texttt{Does [gene] in [indication] have a high expression?} &
\texttt{Does [gene] in [indication] have a high expression in cancer cells?} \\
\texttt{Does [gene] in [indication] have a high expression in cancer cells that also have a high endocytosis signature?} &
\texttt{Does [gene] in [indication] have a high expression in malignant cells from tumours that also have a high cathepsin signature?} \\
\texttt{Does [gene] in [indication] have a high expression in tumours that also have a high cathepsin signature?} &
\texttt{Does [gene] in [indication] have a high proportion of malignant cell pseudobulks with high expression?} \\
\texttt{Does [gene] in [indication] have a high proportion of malignant cells expressing it?} &
\texttt{Does [gene] in [indication] have a high proportion of patients with copy number alterations for this gene?} \\
\texttt{Does [gene] in [indication] have a high proportion of tumours with high expression within at least one cancer indication?} &
\texttt{Does [gene] in [indication] have a high Quasi H score in the pseudobulk of malignant cells?} \\
\texttt{Does [gene] in [indication] have a high spatial autocorrelation of expression?} &
\texttt{Does [gene] in [indication] have a high tumour quasi H score?} \\
\texttt{Does [gene] in [indication] have a higher expression in cancer cells versus all other cells in the tumour?} &
\texttt{Does [gene] in [indication] have a higher expression in tumour versus tumour adjacent normal tissue?} \\
\texttt{Does [gene] in [indication] have a higher expression in tumour core versus tumour edge in spatial data?} &
\texttt{Does [gene] in [indication] have a higher expression in tumour versus blood?} \\
\texttt{Does [gene] in [indication] have a higher expression in tumour versus bone marrow?} &
\texttt{Does [gene] in [indication] have a higher expression in tumour versus healthy tissues?} \\
\texttt{Does [gene] in [indication] have a higher expression in tumour versus heart?} &
\texttt{Does [gene] in [indication] have a higher expression in tumour versus kidney?} \\
\texttt{Does [gene] in [indication] have a higher expression in tumour versus liver?} &
\texttt{Does [gene] in [indication] have a higher expression in tumour versus spleen?} \\
\texttt{Does [gene] in [indication] have a higher expression in tumour versus stroma in spatial data?} &
\texttt{Does [gene] in [indication] have a higher proportion of malignant cells than of immune cells expressing it?} \\
\texttt{Does [gene] in [indication] have a higher proportion of malignant cells than of stromal cells expressing it?} &
\texttt{Does [gene] in [indication] have a low level of heterogeneity in expression levels between malignant cell subclusters?} \\
\texttt{Does [gene] in [indication] have a positive spatial association with cathepsin signature?} &
\texttt{Does [gene] in [indication] have a positive spatial association with endocytosis signature?} \\
\texttt{Does [gene] in [indication] have a spatial expression distribution so that malignant spots not expressing the gene are close neighbors of malignant spots expressing the gene (rather than far away)?} &
\texttt{Does [gene] in [indication] have a homogeneous and stable spatial expression?} \\
\bottomrule
\caption{Full set of question types instantiated from GI feature scores.}
\label{tab:gi_qtypes}
\end{longtable}

\paragraph{Q\&A Schema}
\textit{(Schema examples; not exhaustive)}
\begin{quote}
\ttfamily
Is the \{transcript abundance / mRNA abundance / gene expression level\} of \{GENE\} significantly elevated in \{INDICATION\} tumour tissue compared to \{REFERENCE\_NORMAL\_TISSUE\}?\\
A) True \quad B) False (option ordering shuffled) \hfill \textbf{Answer:} \{A/B\}
\end{quote}
\begin{quote}
\ttfamily
Is the frequency of copy number variations (CNVs) affecting the \{GENE\} gene significantly elevated in \{INDICATION\} patients?\\
A) True \quad B) False (option ordering shuffled) \hfill \textbf{Answer:} \{A/B\}
\end{quote}

\paragraph{Q\&A Examples}
\begin{quote}
\textbf{Q:} Is the gene expression level of PLEKHG6 significantly elevated in bladder urothelial carcinoma tumour tissue compared to normal spleen tissue?\\
A) True B) False\\
\textbf{A:} A
\end{quote}
\begin{quote}
\textbf{Q:} Is the frequency of copy number variations (CNVs) affecting the HOXC8 gene elevated in Bladder urothelial carcinoma patients?\\
A) True B) False\\
\textbf{A:} B
\end{quote}

\paragraph{Train/Test Split Strategy}
We enforce conservative splits to avoid leakage of decision cues:
\begin{itemize}
    \item \textbf{Disjoint gene} across train/test.
    \item \textbf{Stratification} by (i) feature/question type and (ii) label balance (True/False) within indication.
\end{itemize}

% ---------------------------

\subsection{\glsfirst{tcgasa}}

We prepared four datasets from TCGA bulk RNA-seq to test reasoning over gene set (signature) activities: (1) Signature Expression, (2) Signature Similarity, (3) Cancer Similarity, and (4) Cancer Signature Comparison. 

\subsubsection{Signature Expression}
\label{supplementary:datasets-tcgasa-se}
\paragraph{Description}
Questions compare computed signature expression levels across cancer types, where signatures represent averaged activity of curated gene sets. This dataset comprises 4000 training and 800 test samples.

\paragraph{Curation Strategy}
Signatures activities were computed using ssGSEA~\citep{barbie_ssgsea_2009}, which corresponds to the difference of the average ranks of the genes in the gene set and the average ranks of all remaining protein-coding genes. Gene expressions were downloaded from the GDC server using TCGAbiolinks~\citep{mounir2019-pi} and taken as log TPMs. The gene sets were retrieved from the Perturbagen Signatures collection of DSigdb~\citep{yoo_dsigdb_2015}, which lists genes that are significantly differentially expressed for each compound in the Connectivity Map, resulting in 2,000 gene sets covering 11,000 genes. Prompts include the signature name and up to 10 genes of the gene set.

\paragraph{Q\&A Types}
Binary multiple choice (\texttt{signature\_expression\_binary}): given a signature, choose which of two cancer types shows higher average activity of that signature.

\paragraph{Q\&A Schema}
\begin{quote}
\ttfamily
Which cancer type has higher expression of the \{GENESET\_NAME\} (computed as the average activity of: \{GENESET\_GENES\} and \{REMAINING\_GENE\_COUNT\} more genes) signature?\\
A) \{CANCER\_NAME\_A\} \quad B) \{CANCER\_NAME\_B\}\\
\textbf{Answer:} \{A/B\}
\end{quote}

\paragraph{Q\&A Example}
\begin{quote}
\textbf{Q:} Which cancer type has higher expression of the l-thyroxine (computed as the average activity of: ABCB1, AHR, PPP1CA, PLA2G7, PIP4K2A, PPARG, NFE2L2, ATG4B, THRB, NR3C1, and 1 more genes) signature?\\
A) Cholangiocarcinoma\quad B) Prostate adenocarcinoma\\
\textbf{A:} A
\end{quote}

\paragraph{Train/Test Split Strategy}
The subject is the signature. Training and test splits have a disjoint set of subjects, as well as possible candidates to choose the anwser from.
For this dataset we use all 33 TCGA indications with RNAseq data, and keep the following for the test set: ``TGCT'', ``KICH'', ``CESC'', ``READ'', ``LGG'', ``SARC'', ``READ''.
Questions are balanced by the correct answer content label (\texttt{A/B}) and stratified by both the subject and the correct answer content (\texttt{CANCER\_NAME}) within each split.

\subsubsection{Signature Similarity}

\paragraph{Description}
This dataset evaluates distributional similarity between molecular signatures across cancer types using maximum mean discrepancy or sliced Wasserstein distance metrics. Questions assess which signatures show more similar activity patterns to reference signatures, requiring models to reason about molecular pathway similarities and biological mechanism overlaps across different cancer contexts. In total, 970 samples are allocated to the training set and 200 to the test set.

\paragraph{Curation Strategy}
First, signature activities are computed in the same manner as described in~\cref{supplementary:datasets-tcgasa-se}, across all TCGA indications. For each signature pair, the sliced Wasserstein distance is calculated to determine the distributional distance within each cancer indication. This value is then averaged across all indications to establish a global distance measure.

\paragraph{Q\&A Types}
Binary multiple choice (\texttt{signature\_similarity\_binary}): given a reference signature, select which of two candidate signatures has a more similar activity distribution across all cancer types.

\paragraph{Q\&A Schema}
\begin{quote}
\ttfamily
Which signature has a more similar distribution to \{GENESET\_NAME\_REFERENCE\} (computed as the average activity of: \{GENESET\_GENES\_REFERENCE\}, and \{REMAINING\_GENE\_COUNT\_REFERENCE\} more genes) across all cancer types?\\
A) \{GENESET\_NAME\_A\} (computed as the average activity of \{GENESET\_GENES\_A\}, and \{REMAINING\_GENE\_COUNT\_A\} more genes) \quad B) \{GENESET\_NAME\_B\} (computed as the average activity of \{GENESET\_GENES\_B\}, and \{REMAINING\_GENE\_COUNT\_B\} more genes)\\
\textbf{Answer:} \{A/B\}
\end{quote}

\paragraph{Q\&A Example}
\begin{quote}
\textbf{Q:} Which signature has a more similar distribution to ethinyl\_estradiol (computed as the average activity of: SLC22A2, NLRP3, NLRP1, CXCL8, TP53, CYP2C19, CYP2D6, OPRK1, UGT1A1, AR, and 27 more genes) across all cancer types?\\
A) apomorphine (computed as the average activity of: AHR, HTR2C, EHMT2, DRD2, HTR1A, HTR2A, AR, HSD17B10, DRD4, MAPT, and 11 more genes) \quad B) entacapone (computed as the average activity of: COMT, HSPB1, NFE2L2, POLK, TP53, UGT1A9)\\
\textbf{A:} A
\end{quote}

\paragraph{Train/Test Split Strategy}
The subject is the reference signature. In both splits, the reference and both candidate signatures are drawn exclusively from the corresponding partition, creating completely disjoint similarity matrices between train and test.
Questions are stratified by the correct answer content label (\texttt{A/B}), by the subject (\texttt{GENESET\_NAME\_REFERENCE}) and by the correct answer content (\texttt{GENESET\_NAME}) within each split.

\subsubsection{Cancer Similarity}

\paragraph{Description}
The largest individual dataset in the task collection (30,000 training and 400 test samples), this dataset evaluates, for a given reference cancer type and a fixed signature, which of two candidate cancer types has a signature-activity distribution more similar to the reference cancer.

\paragraph{Curation Strategy}
For each signature, we compute activity distributions for all cancer types. Using a distance metric (Sliced Wasserstein or MMD), we compare the subject cancer to two candidates and ask which is closer. Prompts include the signature name and its gene list snippet to anchor biological context.

\paragraph{Q\&A Types}
One binary comparison task:
\begin{itemize}
    \item \texttt{cancer\_similarity\_binary}: which of the two cancer types is more similar to reference cancer type based on a signature activity?
\end{itemize}

\paragraph{Q\&A Schema}
\begin{quote}
\ttfamily
Based on \{GENESET\_NAME\} (computed as the average activity of: \{GENESET\_GENES\}) signature activity patterns from bulk RNA-seq data, which cancer type is more similar to \{CANCER\_NAME\_REFERENCE\}?\\
A) \{CANCER\_NAME\_A\} \quad B) \{CANCER\_NAME\_B\}\\
\textbf{Answer:} \{A/B\}
\end{quote}

\paragraph{Q\&A Example}
\begin{quote}
\textbf{Q:} Based on methylenediphosphonic\_acid (computed as the average activity of: BAZ2B, EHMT2, KDM4E, PPP1CA, PTBP1) signature activity patterns from bulk RNA-seq data, which cancer type is more similar to Bladder urothelial carcinoma?\\
A) Stomach adenocarcinoma \quad B) Pancreatic adenocarcinoma\\
\textbf{A:} A
\end{quote} 

\paragraph{Train/Test Split Strategy}
The subject is the reference cancer type. In both splits, the subject cancer type and the two candidate cancer types are drawn from the corresponding partition, creating a disjoint similarity matrix between the two sets.
The test set is composed of the following indications: ``THCA'', ``GBM'', ``CESC'', ``PRAD'', ``SKCM'', ``UCS'', ``UVM'', ``ACC'', ``PCPG'', ``KIRC'', ``OV'', ``ESCA'', ``UCEC'', ``LGG'', ``LUAD'', ``SARC''.
Questions are stratified by the correct answer content label (\texttt{A/B}), by the subject (\texttt{CANCER\_NAME\_REFERENCE}) and by the correct answer content (\texttt{CANCER\_NAME}) within each split.

\subsubsection{Cancer Signature Comparison}

\paragraph{Description}
Within a single cancer type, asks which of two signatures shows higher activity. The dataset contains 4000 training and 800 test samples.

\paragraph{Curation Strategy}
For each cancer, we rank signatures by activity and sample a pair from high vs. low tail to ensure a discriminative pair. Prompts include the cancer's full name and both signatures with gene set snippets as options.

\paragraph{Q\&A Types}
One binary comparison task:
\begin{itemize}
    \item \texttt{cancer\_signatures\_comparison}: which of the two signatures show a higher expression in the cancer of interest?
\end{itemize}

\paragraph{Q\&A Schema}
\begin{quote}
\ttfamily
In \{CANCER\_NAME\} which signature has higher expression?\\
A) \{GENESET\_NAME\_A\} (computed as the average activity of \{GENESET\_GENES\_A\}) \quad B) \{GENESET\_NAME\_B\} (computed as the average activity of \{GENESET\_GENES\_B\})\\
\textbf{Answer:} \{A/B\}
\end{quote}

\paragraph{Q\&A Example}
\begin{quote}
\textbf{Q:} In Cholangiocarcinoma, which signature has higher expression?\\
A) enalapril\_maleate (computed as the average activity of: ABCB1, ACE, ARRB1, INS, KDM4A) \quad B) nalbuphine (computed as the average activity of: CYP1A2, CYP2D6, CYP3A4, OPRD1, OPRK1, OPRM1)\\
\textbf{A:} A
\end{quote} 

\paragraph{Train/Test Split Strategy}
The subject is the cancer type. In both the training and test splits, the subject cancer type and the two candidate signatures are drawn from the corresponding partitions.
For this dataset we use all 33 TCGA indications with RNAseq data, and keep the following for the test set: ``READ'', ``SARC'', ``TGCT'', ``CESC'', ``KICH'', ``LGG''.
Questions are stratified by the correct answer content label (\texttt{A/B}), by the subject (\texttt{CANCER\_NAME}) and by the correct answer content (\texttt{GENESET\_NAME}) within each split.

% ---------------------------

\subsection{\glsfirst{dseqde}}
\paragraph{Description}
The \gls{dseqde} dataset probes whether inhibiting a specific target leads to transcriptional deregulation of a candidate readout in a defined cancer context. Each item is framed as a natural-language multiple-choice question. We include (i) binary Yes/No questions of the form ``does inhibiting target~$T$ deregulate gene~$g$?'', (ii) pairwise gene comparisons ``which of these two genes is deregulated when inhibiting~$T$?'', and (iii) an ablation variant at the pathway level using Reactome, asking for the affected pathway. The task emphasises \emph{target\,\textrightarrow\,response} reasoning rather than compound memorisation.

\paragraph{Curation Strategy}
We built the QA pairs from proprietary drug perturbation assay data. For each compound treatment with matched controls, we:
\begin{enumerate}
  \item \textbf{Target mapping:} associate compounds to their annotated inhibitory target(s) using curated compound metadata; compounds without a clear inhibitory mechanism or with missing target annotations are excluded.
  \item \textbf{Differential expression (DEA):} compute treated vs.\ control contrasts to obtain sets of differentially expressed genes (DEGs) per target.
  \item \textbf{Filtering to inhibitors:} retain only compounds acting as inhibitors (including ATP-competitive, allosteric, covalent inhibitors, antagonists, degraders, etc.) to ensure a consistent \emph{loss-of-function} interpretation.
  \item \textbf{Balancing:} to control class imbalance, we downsample non-DEGs when forming negatives so that, for each target, the number of negative gene items matches the positives.
  \item \textbf{Question generation:} produce natural-language items with randomized A/B answer assignment; metadata store the target (and gene/pathway when relevant). Pairwise items sample one deregulated feature and one non-deregulated feature from the same candidate universe. Reactome variants are produced by mapping DEGs to pathway gene sets.
\end{enumerate}

\paragraph{Q\&A Types}
\begin{itemize}
  \item \texttt{Yes/No (gene level)}: ``Would a drug inhibiting the activity of a target induce a deregulation of a gene in given cancer cells?''
  \item \texttt{Pairwise (gene level)}: ``Which of these two genes would be deregulated by a drug inhibiting given target in given cancer cells?''
  \item \texttt{Pairwise (Reactome pathway ablation)}: ``Which of these two pathways would be deregulated by a drug inhibiting the activity of a target in given cancer cells?''
\end{itemize}

\paragraph{Q\&A Schema}
\begin{quote}\ttfamily
Would a drug inhibiting the activity of the target \{TARGET\} induce a deregulation of gene \{GENE\} in \{CANCER\_TYPE\} cells?\\
A) Yes \quad B) No (option ordering shuffled) \hfill \textbf{Answer:} \{A/B\}
\end{quote}
\begin{quote}\ttfamily
Which of these two genes would be deregulated by a drug inhibiting the activity of the target \{TARGET\} in \{CANCER\_TYPE\} cells?\\
A) \{GENE\_A\} \quad B) \{GENE\_B\}\hfill \textbf{Answer:} \{A/B\}
\end{quote}
\begin{quote}\ttfamily
Which of these two pathways would be deregulated by a drug inhibiting the activity of the target \{TARGET\} in \{CANCER\_TYPE\} cells?\\
A) \{PATHWAY\_A\} \quad B) \{PATHWAY\_B\}\hfill \textbf{Answer:} \{A/B\}
\end{quote}

\paragraph{Q\&A Examples}
\begin{quote}
\textbf{Q:} Would a drug inhibiting the activity of the target PIK3CA induce a deregulation of gene UBL3 in muscle invasive bladder cancer cells?\\
A) Yes \quad B) No\\ 
\textbf{A:} B
\end{quote}
\begin{quote}
\textbf{Q:} Which of these two genes would be deregulated by a drug inhibiting the activity of the target PIK3CD in muscle invasive bladder cancer cells?\\
A) TNFRSF19 \quad B) TRAF7\\
\textbf{A:} A
\end{quote}
\begin{quote}
\textbf{Q:} Which of these two pathways would be deregulated by a drug inhibiting the activity of the target CDK9 in muscle invasive bladder cancer cells?\\
A) Degradation of GLI1 by the proteasome \quad B) Signaling by EGFRvIII in Cancer\\
\textbf{A:} B
\end{quote}

\paragraph{Train/Test Split Strategy}
We constructed a conservative split to prevent information leakage:
\begin{itemize}
\item No overlap of subject entities between splits, where a ``subject'' is defined as any \texttt{target}, \texttt{gene} (in the question or answer options), or \texttt{pathway} (in the answer options).
\item For each target, the number of positive (deregulated) and negative (non-deregulated) examples is kept equal in both the training and test sets.
\item The A/B answer option order is randomized independently within each split.
\end{itemize}

% ---------------------------

\subsection{\glsfirst{dpp}}

\paragraph{Description}  
The \gls{dpp} dataset is derived from the Tahoe-100M single-cell perturbation screen~\citep{zhang_tahoe-100m_2025}, the largest transcriptomic perturbation dataset to date (at the time of writing), measuring the effects of 1,100 small-molecule perturbations across 50 cancer cell lines. Our Q\&A pairs here, focus on Reactome pathway-level differential expression analysis, using ssGSEA to identify the \textbf{most perturbed pathway} in each (drug, cell line, concentration) context.  
``Most perturbed'' here is defined as the significantly enriched gene set with the largest absolute Normalized Enrichment Score (NES).

\paragraph{Curation Strategy}  
The raw Tahoe-100M profiles are stored as plate-level \texttt{.h5ad} files in public Google Cloud storage.  
Our processing pipeline was as follows:
\begin{enumerate}
    \item Load plate-level expression matrices into \texttt{AnnData} objects using \texttt{scanpy}.
    \item For each (drug, cell line, concentration) context, rank genes by differential expression (treated vs.\ control) using \texttt{rank\_genes\_groups}.
    \item Run ssGSEA with Reactome gene sets to compute enrichment scores per context. Keep gene sets with robust enrichment results (FDR~<~0.05).
    \item Identify the pathway with the largest absolute NES whilst recording its direction of deregulation (upregulated / downregulated).
    \item Map these pathway calls into natural-language multiple-choice \gls{qa} format, with two answer options (A or B).
\end{enumerate}

\paragraph{\gls{qa} Types}  
The dataset contains a single question type:
\begin{itemize}
    \item \texttt{most\_perturbed\_pathway}: Identify the Reactome pathway most significantly affected by a given drug treatment in a specific cell line at a specified concentration, including the direction of change.
\end{itemize}

\paragraph{\gls{qa} Schema}

\begin{quote}
\ttfamily
Which Reactome gene set would be most significantly affected by \{DRUG\} at \{DRUG\_CONC\} µM in \{CELL\_LINE\} cells, and in which direction: upregulation or downregulation?\\
A) \{PATHWAY\_A\} - \{DIRECTION\_A\} \quad B) \{PATHWAY\_B\} - \{DIRECTION\_B\}\\
\textbf{Answer:} \{A/B\}
\end{quote}

Each record contains the subjects:
\begin{itemize}
    \item \texttt{DRUG}: Name of the perturbing compound
    \item \texttt{DRUG\_CONC}: Drug concentration in µM
    \item \texttt{CELL\_LINE}: Name of the cell line used in the experiment
    \item choices: Two possible \texttt{PATHWAY - DIRECTION} options (A/B)
\end{itemize}

\paragraph{Q\&A Example}  
\begin{quote}
\textbf{Q:} Which Reactome gene set would be most significantly affected by Saquinavir in A549 cells at 0.05~µM, and in which direction: upregulation or downregulation? \\
A) Nuclear pore complex (NPC) disassembly - downregulated \\
B) SARS-CoV-1 modulates host translation machinery - upregulated \\
\textbf{A:} B
\end{quote}

\paragraph{Train/Test Split Strategy}  
We designed the split to prevent leakage across three key dimensions: compounds, cell lines, and Reactome gene sets.

\begin{enumerate}
    \item \textbf{No shared compounds} between train and test sets.
    \item \textbf{No shared cell lines} between train and test sets.
    \item \textbf{No shared or highly similar Reactome gene sets:}  
    \begin{enumerate}
        \item \textbf{Build the Reactome ontology.}  
        We used the official Reactome pathways relations~\citep{reactome_pathways_relation} to construct a directed graph, where:
        \begin{itemize}
            \item Each node corresponds to a gene set (with a stable Reactome ID).
            \item Each edge encodes a hierarchical ``parent--child'' relationship between pathways, where the child pathway is a more specific subdivision of the parent pathway.  
        \end{itemize}
        This yields a tree-like hierarchy of pathway modules.
        
        \item \textbf{Identify subtrees (functional modules).}
        From this graph, we:
        \begin{itemize}
            \item Identify roots (top-level categories like ``Signal Transduction'', ``Immune System'')
            \item For each root, extract its full subtree of descendants. Each subtree is hence a functional module.
        \end{itemize}
        This groups gene sets into biological units where gene sets are semantically and biologically related.
        
        \item \textbf{Assign subtrees alternately to train and test.}  
        We sort subtrees by size and assign them alternately to train and test. This ensures non-overlapping biological families between splits. E.g., if ``Immune System'' goes to train, then ``Cell Cycle'' may go to test.
        
        \item \textbf{Filter leaky test pathways by Jaccard similarity.}  
        Compute the maximum Jaccard gene overlap between each candidate test pathway and all train pathways. Retain only the test pathways with similarity $\leq 0.3$; discard the rest.
    \end{enumerate}
\end{enumerate}

This splitting strategy guarantees that the train and test sets are disjoint with respect to \textbf{compounds}, \textbf{cell lines}, and \textbf{gene sets}, requiring models to generalise simultaneously to unseen perturbations, unseen biological contexts, and unseen pathway modules.

% \begin{quote}\ttfamily
% Training gene set term count: \phantom{0}643 \\
% Test gene set term count: \phantom{00}421 \\
% Discarded test sets: \phantom{00}183 \\
% Max similarity between train and test gene sets: 0.2986111111
% \end{quote}

% ---------------------------

\subsection{\glsfirst{ttp}}
\paragraph{Description}
Multi-domain True/False (Yes/No) questions assessing target \emph{druggability}, \emph{preferred modality} (small molecule vs.\ antibody), \emph{structural characterisation}, \emph{ligand knowledge}, \emph{safety/toxicity}, \emph{inflammatory/immunological involvement}, and \emph{cancer biology} relevance.
Items are multiple-choice A/B selections, phrased in formal biomedical language with alternative and negative phrasings to increase linguistic variety.

\paragraph{Curation Strategy}
Source data are an aggregation of multiple knowledge sources (\emph{UniProt}, patent databases, and clinical trial (CT) databases). We prepared decision tables for small-molecule and antibody tractability (e.g., fields \texttt{decision\_sm} and \texttt{decision\_ab} with rationales such as ``CT'', ``other'', ``not accessible'', extracellular region size, etc.) and transformed them into Q\&A pairs using a reproducible template. We instantiated binary prompts such as \emph{``Can \{GENE\} be targeted by a small molecule?''} or \emph{``Can \{GENE\} be targeted by an antibody?''} with options \{A,B\}=\{Yes,No\} or \{No,Yes\}, plus additional domains (\emph{structure}, \emph{ligand}, \emph{toxicity}, \emph{inflammatory/immunological}, \emph{cancer biology}, \emph{general/modality}). Alternative wordings (suffix \texttt{\_alt}) and deliberately flipped/negative items (suffix \texttt{\_negative}) are included.

Each record stores provenance in \texttt{metadata}, e.g.\ \{\texttt{target\_protein}, \texttt{original\_question}, \texttt{original\_answer}, \texttt{answer\_type} (binary/categorical), \texttt{question\_category}, \texttt{template\_used}, \texttt{data\_row\_index}, optional \texttt{is\_alternative\_phrasing}, \texttt{is\_negative\_example}, \texttt{original\_phrasing}\}.

\paragraph{Q\&A Types}
We include the following \texttt{question\_type} categories (counts are totals across train and test; percentages are relative to the total dataset size of 2{,}758 items):

\begin{center}
\small
\begin{tabular}{l r r@{\hspace{1.5em}} l r r}
\toprule
Type & Count & \% & Type & Count & \% \\
\midrule
\texttt{antibody} & 96 & 3.48 & \texttt{ligand} & 48 & 1.74 \\
\texttt{antibody\_alt} & 25 & 0.91 & \texttt{ligand\_alt} & 10 & 0.36 \\
\texttt{antibody\_negative} & 17 & 0.62 & \texttt{ligand\_negative} & 10 & 0.36 \\
\texttt{cancer\_biology} & 48 & 1.74 & \texttt{multiple\_choice} & 2{,}000 & 72.51 \\
\texttt{cancer\_biology\_alt} & 17 & 0.62 & \texttt{small\_molecule} & 144 & 5.22 \\
\texttt{cancer\_biology\_negative} & 9 & 0.33 & \texttt{small\_molecule\_alt} & 42 & 1.52 \\
\texttt{druggability} & 45 & 1.63 & \texttt{small\_molecule\_negative} & 22 & 0.80 \\
\texttt{druggability\_alt} & 13 & 0.47 & \texttt{structure} & 45 & 1.63 \\
\texttt{druggability\_negative} & 11 & 0.40 & \texttt{structure\_alt} & 13 & 0.47 \\
\texttt{general} & 26 & 0.94 & \texttt{structure\_negative} & 10 & 0.36 \\
\texttt{general\_alt} & 9 & 0.33 & \texttt{toxicity} & 15 & 0.54 \\
\texttt{general\_negative} & 5 & 0.18 & \texttt{toxicity\_alt} & 3 & 0.11 \\
\texttt{inflammatory\_immunological} & 45 & 1.63 & \texttt{toxicity\_negative} & 4 & 0.14 \\
\texttt{inflammatory\_immunological\_alt} & 14 & 0.51 & & & \\
\texttt{inflammatory\_immunological\_negative} & 12 & 0.44 & & & \\
\bottomrule
\end{tabular}
\end{center}

\paragraph{Q\&A Schema}
All items are binary with two options:
\begin{quote}\ttfamily
Is/Can/Does \{TARGET\} \{predicate\}?\\
A) Yes \quad B) No (option ordering shuffled) \hfill \textbf{Answer:} \{A/B\}
\end{quote}
Representative predicates include: \emph{druggable}; \emph{suitable for small molecule development}; \emph{suitable for antibody development}; \emph{has a known ligand}; \emph{has been structurally characterised}; \emph{linked to toxicity issues / safety concerns}; \emph{involved in inflammatory diseases}; \emph{associated with cancer pathways}; as well as more general \emph{modality} prompts.

\paragraph{Q\&A Examples}
\begin{quote}
\textbf{Q:} Can TEX46 be targeted by a small molecule?\\
A) no \quad B) yes \hfill \textbf{A:} A \ (\texttt{multiple\_choice})
\end{quote}
\begin{quote}
\textbf{Q:} Is IL\textendash28 druggable?\\
A) No \quad B) Yes \hfill \textbf{A:} A \ (\texttt{druggability})
\end{quote}
\begin{quote}
\textbf{Q:} Can PRDX5 be targeted by antibodies?\\
A) No \quad B) Yes \hfill \textbf{A:} A \ (\texttt{antibody})
\end{quote}
\begin{quote}
\textbf{Q:} Has PD\textendash1 been structurally characterized?\\
A) Yes \quad B) No \hfill \textbf{A:} A \ (\texttt{structure})
\end{quote}
\begin{quote}
\textbf{Q:} Is VEGF suitable for small molecule development?\\
A) Yes \quad B) No \hfill \textbf{A:} A \ (\texttt{small\_molecule})
\end{quote}
\begin{quote}
\textbf{Q:} Does ITGA3 have a known ligand?\\
A) No \quad B) Yes \hfill \textbf{A:} B \ (\texttt{ligand})
\end{quote}
\begin{quote}
\textbf{Q:} Is KRAS linked to toxicity issues?\\
A) Yes \quad B) No \hfill \textbf{A:} A \ (\texttt{toxicity})
\end{quote}
\begin{quote}
\textbf{Q:} Is TIGIT associated with cancer pathways?\\
A) Yes \quad B) No \hfill \textbf{A:} A \ (\texttt{cancer\_biology})
\end{quote}

\paragraph{Train/Test Split Strategy}
Splits were produced via a simple random split without additional subject (\emph{target}) disjoint constraints. 
Targets (genes) can appear multiple times within the same split under different \texttt{question\_type}s and/or phrasings, and the same targets as well as full questions can occur in both train and test; thus the splits are not subject-disjoint. 
Additionally there are exact duplicate questions within the train set and within the test set, and cases where the exact same question text appears with different correct answers due to conflicting label assignments across variants (e.g., multiple occurrences of ``Can NIK be targeted by small molecules?'' with both \texttt{A} and \texttt{B} as correct answers under different \texttt{question\_type} labels). 

% ---------------------------

\subsection{\glsfirst{sd}}

\paragraph{Description}
The \gls{sd} dataset evaluates pairwise comparison of predicted pocket druggability within a single protein. Each item asks which of two candidate binding sites (enumerated as lists of residues from the protein's original sequence, using one-letter amino-acid codes with sequence indices) has the higher druggability score. Protein structures are sourced from experimentally solved datasets (TOUGH-M1~\citep{tough-m1_govindaraj_2018}), and pocket identification/scoring is performed with \texttt{Fpocket}~\citep{le_guilloux_fpocket_2009}.

\paragraph{Curation Strategy}
Binding sites (pockets) and their druggability scores are computed with \texttt{Fpocket} on all protein structures from the TOUGH-M1 dataset. For each protein, provided as a sequence of residues (no 3D coordinates), two candidate binding sites are presented in the form of their corresponding list of residues with respect to the original sequence; one of them is the pocket with highest druggability score (the correct answer), the other one is randomly chosen among the remaining pockets. The two candidate pockets are randomly presented as either pocket A or pocket B. 
Please note that the druggability score involves in particular structural features of the protein, but no structural information is provided to the algorithm directly, only sequential data.

\paragraph{Q\&A Types}
Single binary comparison task:
\begin{itemize}
    \item \texttt{druggability}: given two candidate binding sites on the same protein, select the site with the higher druggability score.
\end{itemize}

\paragraph{Q\&A Schema}
\begin{quote}
\ttfamily
Given the protein with amino-acid sequence \{SEQUENCE\}, which one of these two binding sites has the highest druggability score?\\
A) \{RESIDUES\_A\} \quad B) \{RESIDUES\_B\}\\
\textbf{Answer:} \{A/B\}
\end{quote}

\paragraph{Q\&A Examples}
\begin{quote}
\textbf{Q:} Given the protein with amino-acid sequence M1 I2 T3 C4 G5 Q6 V7 S8 S9 S10 L11 A12 P13 C14 I15 P16 Y17 V18 R19 G20 G21 G22 A23 V24 P25 P26 A27 C28 C29 N30 G31 I32 R33 N34 V35 N36 N37 L38 A39 R40 T41 T42 P43 D44 R45 Q46 A47 A48 C49 N50 C51 L52 K53 Q54 L55 S56 A57 S58 V59 P60 G61 V62 N63 P64 N65 N66 A67 A68 A69 L70 P71 G72 K73 C74 G75 V76 S77 I78 P79 Y80 K81 I82 S83 A84 S85 T86 N87 C88 A89 T90 V91 K92, which one of these two binding sites (specified by the corresponding amino-acids from the original sequence) has the highest druggability score?\\
A) P71 L70 A67 I82 I78 V62 S56 V18 L55 I15 S8 V59 L11 C14 I32 V35 A12 L52 \quad B) A68 I78 P71 Y80 I82 K81 P79 G72 V76\\
\textbf{A:} A
\end{quote}

\begin{quote}
\textbf{Q:} Given the protein with amino-acid sequence E1 A2 T3 K4 A5 R6 I7 F8 E9 A10 A11 V12 A13 E14 F15 A16 R17 H18 G19 I20 A21 G22 A23 R24 I25 D26 R27 I28 A29 A30 E31 A32 R33 A34 N35 K36 Q37 L38 I39 Y40 A41 Y42 Y43 G44 N45 K46 G47 E48 L49 F50 A51 S52 V53 L54 E55 K56 K57 L58 D59 L60 A61 I62 S63 V64 P65 V66 D67 P68 D69 D70 I71 E72 G73 W74 I75 D76 R77 L78 L79 D80 Y81 H82 A83 A84 H85 P86 E87 L88 L89 R90 L91 L92 F93 W94 E95 G96 E97 Y98 G99 T100 A101 E102 L103 P104 H105 E106 A107 E108 R109 Q110 E111 H112 Y113 A114 R115 K116 V117 A118 A119 V120 R121 D122 G123 Q124 E125 R126 G127 V128 I129 T130 D131 A132 I133 P134 A135 P136 D137 L138 L139 F140 L141 L142 V143 A144 A145 N146 W147 A148 V149 V150 V151 P152 Q153 K154 R155 I156 L157 V158 G159 G160 G161 D162 A163 G164 T165 D166 G167 L168 R169 D170 S171 I172 K173 K174 A175 A176 R177 R178 I179 V180 D181 R182, which one of these two binding sites (specified by the corresponding amino-acids from the original sequence) has the highest druggability score?\\
A) E87 P86 R90 A16 \quad B) L92 R109 L54 L88 E55 E108 H112 L60 N146 L89 K116 H82 L78 L58 K57 Y113 P136 V117 F140 V143 A61 L139\\
\textbf{A:} B
\end{quote}

\paragraph{Train/Test Split Strategy}
The subject is the protein sequence. Train and test sets have a disjoint set of subjects.
Each question has unique options given its subject, so options are also disjoint between train and test.
Questions are stratified by the correct answer content label (\texttt{A/B}).
% ---------------------------

\section{Detailed Judge prompt for faithfulness analysis}%
\label{supplementary:judge-prompt}
\begin{quote}
    \begin{lstlisting}[basicstyle=\small\ttfamily,breaklines=true]
<|im_start|>system
You are evaluating two responses to a biology question. Your task is to assess which response demonstrates better biological reasoning.


Evaluate the responses based on these specific criteria:
1. Scientific accuracy
2. Logical coherence and depth of reasoning
3. Relevance and completeness of explanation
4. Clarity and precision of language


Then provide your final rating:
- If Response 1 is better: +1
- If Response 2 is better: -1


You MUST respond by first justifying your rating, then a JSON object in this exact format:
<json>
{{"rating": <rating>}}
</json>


Where <rating> is -1 or 1.
<|im_end|>


<|im_start|>user
Question: "{question}"


<Response 1>
{response1}
</Response 1>


<Response 2>
{response2}
</Response 2>
<|im_end|>


<|im_start|>assistant

\end{lstlisting}
\end{quote}

%%%%%%%%%%%%%%%%%%%%%%%%%%%%%%%%%%%%%%%%%%%%%%%%%%%%%%%%%%%%

\end{document}